\documentclass{article}

\usepackage{arxiv}

\usepackage[utf8]{inputenc} 
\usepackage[T1]{fontenc}    
\usepackage{hyperref}       
\usepackage{url}            
\usepackage{booktabs}       
\usepackage{amsfonts}       
\usepackage{nicefrac}       
\usepackage{microtype}      
\usepackage{lipsum}		
\usepackage{graphicx}
\usepackage{natbib}
\usepackage{doi}
\usepackage{amsmath,amsfonts,amssymb}

\usepackage{algorithmic}
\usepackage{algorithm}
\newtheorem{definition}{Definition}
\newtheorem{prop}{Proposition}

\usepackage{array}
\usepackage[caption=false,font=normalsize,labelfont=sf,textfont=sf]{subfig}
\usepackage{textcomp}
\usepackage{stfloats}
\usepackage{url}
\usepackage{verbatim}
\usepackage{graphicx}
\usepackage{booktabs, multirow} 
\usepackage{soul}
\usepackage[table]{xcolor} 
\usepackage{changepage,threeparttable} 

\usepackage{relsize} 
\usepackage{amssymb}
\usepackage{array}

\usepackage{latexsym}
\usepackage{amsmath}
\usepackage{url}
\definecolor{newcolor}{rgb}{.8,.349,.1}

\title{Image contrast enhancement based on the Schr\"odinger operator spectrum}

\author{ \href{https://orcid.org/my-orcid?orcid=0000-0001-8100-2822}{\includegraphics[scale=0.06]{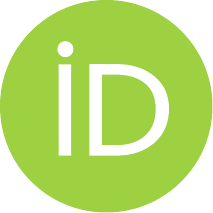}\hspace{1mm}Juan M. Vargas} \\
	National Institute for Research in Digital Science and Technology (INRIA)\\
	Palaiseau, France\\
	\texttt{juan-manuel.vargas-garcia@inria.fr} \\
	\And
	\href{https://orcid.org/0000-0001-5944-0121}{\includegraphics[scale=0.06]{orcid.pdf}\hspace{1mm} Taous-Meriem Laleg-Kirati} \\
	National Institute for Research in Digital Science and Technology (INRIA)\\
	Palaiseau, France \\
	\texttt{Taous-Meriem.Laleg@inria.fr} \\
}

\renewcommand{\shorttitle}{Image contrast enhancement based on the Schr\"odinger operator spectrum}

\hypersetup{
pdftitle={Image contrast enhancement based on the Schr\"odinger operator spectrum},
pdfsubject={cs.CV},
pdfauthor={Juan M. Vargas, Taous-Meriem Laleg-Kirati},
pdfkeywords={Contrast enhancement,  semi-classical signal analysis, Schr\"odinger operator spectrum},
}

\begin{document}
\maketitle

\begin{abstract}
	In this study, we propose a novel image contrast enhancement method based on projecting images onto the squared eigenfunctions of the two-dimensional Schrödinger operator. This projection relies on a design parameter, $\gamma$, which controls pixel intensity during image reconstruction. The method's performance is evaluated using color images. The selection of $\gamma$ values is guided by priors based on fuzzy logic and clustering, preserving the spatial adjacency information of the image. Additionally, multi-objective optimization using the Non-dominated Sorting Genetic Algorithm II (NSGA-II) is employed to determine the optimal values of $\gamma$ and the semi-classical parameter, $h$, from the 2D-SCSA. Results demonstrate that the proposed method effectively enhances image contrast while preserving the inherent characteristics of the original image, producing the desired enhancement with minimal artifacts.\\
\end{abstract}


\section{Introduction}\label{intro}
\noindent  In modern image processing, low-contrast images represent one of the most recurrent problems faced. This problem occurs when the objects of an image present similar pixel values, making it more difficult to interpret the image \cite{VIJA}. Low-contrast images can be generated due to different phenomena and conditions. For example, when images are captured under unstable lighting conditions where the image can present pixels with values too dark or too bright, affecting the interpretability of the image by reducing the contrast and adding noise \cite{Unconds}. Finally, one of the most important cases is medical images, where low-contrast noise images are produced by many sources such as the acquisition practices, patient position, or health condition, leading to several challenges in interpreting the images and increasing the probability of inaccurate diagnosis \cite{medical}.\\
 
\noindent Contrast enhancement methods have as main goal to increase the visual quality and interpretability of the images by increasing the difference between the pixel values of the different objects in the image\cite{VIJA,Unconds}. However, these methods may suffer from different challenges that difficult the performance, for example, finding the appropriate balance between enhancing the contrast of the image and preserving the global characteristics of the original image with the idea of not introducing artifacts or generating unrealistic results \cite{Srinivas, Ari}. Furthermore, contrast enhancement methods may suffer from noise amplification, overshooting, and color mismatch \cite{SALEEM2017161}.\\

\noindent  In recent years, several methods have been proposed to enhance the contrast of images to increase the interpretation and the quality of images. These methods can be divided into non-model-driven, model-driven, and data-driven approaches. Non-model-driven methods, such as histogram-based and fusion-based approaches, have the main goal of redistributing the image's pixel values to enhance the quality of the image\cite{underW}. For example, in 2024, Radman and Chandra \cite{RAHMAN}, proposed a method based on a Tripartite sub-image histogram approach applied for slightly low-contrast natural images to improve the contrast more locally than other traditional histogram-based approaches. However, these methods generate a loss in the texture of the images as well as present a tendency to over-enhance the image \cite{RAHMAN,medical}. Model-based methods use mathematical models such as Retinex, Human visible system models, or statistical and spatial priors to enhance the contrast of the images while also providing high interpretability that allows to gain more insight into the model's performance\cite{underW,Unconds}. For instance,  in 2022,  Hum et al. \cite{Unconds} proposed a contrast enhancement method based on a human visible system model combined with just noticeable differences applied to uncontrolled environments on just noticeable difference, showing a great performance under different lighting conditions. However, these methods tend to present some limitations due to the high computational cost and, in some cases, present some over-enhancement, producing nonrealistic results.\\

\noindent Finally, with increased development in machine and deep learning methods, data-driven methods have recently gained much attention in various computer vision fields\cite{revlli}. The main idea of these methods is to use a learning process to find how to transform a low-contrast image into a high-contrast image. These methods can be fully data-driven, where the image is input to the neural network architecture, such as CNN or GAN, and the output will give a high-contrast image. In contrast, the hybrid approach combines the neural network with other image processing methods or dynamical models to find optimal parameters and obtain the high-contrast image output. For example, in 2023, a novel hybrid approach was proposed by Li et al. in \cite{PWGC}, where the main idea is to blend the traditional technique of gamma transformation using gamma maps with the capabilities of deep learning. This methodology allows them to obtain more details and robust results. However, gamma correction maps present a potential risk of over-exposing.\\

\noindent In this study, we propose a novel image contrast enhancement method based on the two-dimensional semi-classical signal analysis (2D-SCSA) method used in the past for different tasks, such as image denoising, feature extraction, and image reconstruction. The proposed method is referred to as $\gamma$-SCSA  as it explores the effect of parameter $\gamma$ used in the 2D-SCSA to modify the intensity of the pixels and, in combination with other parameters, could reduce noise artifacts simultaneously, as has been shown in previous studies  \cite{Kaisserli2014, Chahid2018}. The main idea of 2D-SCSA is to decompose an image into the squared eigenfunctions of the Schr\"odinger operator whose potential is the image \cite{Kaisserli2014,Laleg-Kirati2013}. This decomposition offers a new way of reconstructing an image and has been used in many applications as a feature extraction method \cite{Vargas} and for denoising images \cite{Kaisserli2014,Chahid2018}. However, this decomposition depends on a design parameter $h$ called the semi-classical parameter that controls the number of eigenvalues used in the reconstruction and parameter $\gamma$, which was set to fixed values in previous studies. This project demonstrates that $\gamma$ plays a crucial role in 2D-SCSA and opens new promising directions for image contrast enhancement. The main advantage of the proposed approach is its ability to reconstruct and filter noisy data, thereby avoiding noise amplification while enhancing important information in the image. In the present research, the effectiveness of $\gamma$-SCSA in performing contrast enhancement is tested in various scenarios, such as low-light, underwater, and medical images. Performance analysis is also performed, and the proposed method is compared with state-of-the-art approaches.\\ 

\section{Two-dimensional semi-classical signal analysis background}

\noindent In this section, we present an overview of the fundamentals and properties of 2D-SCSA, which are later extended to contrast enhancement (Section \ref{contrast scsa}). As mentioned in Section \ref{intro}, 2D-SCSA decomposes an image into the squared eigenfunctions of the Schr\"odinger operator  \cite{Kaisserli2014}. This method has been extended from one dimensional (1D) SCSA proposed by Laleg-Kirati et al. \cite{Laleg-Kirati2013}, which has demonstrated great potential in pulse-shaped signal processing and characterization\cite{li,evan,validation}. 2D-SCSA has been successfully employed for image reconstruction  \cite{Kaisserli2014}, denoising \cite{abderrazak2, Chahid2018}, and feature extraction \cite{Garcia,Vargas}.

\subsection{Definition}
\noindent The 2D-SCSA representation is defined as follows: 
 \begin{definition}
\noindent Let $I(x,y)$ be a positive real-valued function, the image representation  $I_{h,\gamma}(x,y)$ of $I(x,y)$ using 2D-SCSA is defined as follows:
\begin{equation}
    I_{h,\gamma}(x,y)= \left [ \dfrac{h^{2}}{L_{\gamma}^{cl}}\sum_{m=1}^{M_{h}} ( -\lambda_{mh})^{\gamma} \Psi_{mh}^{2}(x,y) \right ]^{\frac{1}{\gamma + 1}}
\end{equation}
\noindent where $h$ $\in \mathbb{R}_{+}^{*}$ is the semi-classical signal parameter, $\gamma \in \mathbb{R}_{+}$  , $\lambda_{mh}$ represents the negative eigenvalues, and ${\Psi_{1h}, \Psi_{2h},\dots,\Psi_{M_h}}$ correspond to their associated $L^{2}$-normalized eigenfunctions. $M_{h}$ is the number of negative eigenvalues extracted from the two-dimensional semi-classical Schr\"odinger operator, which is described as follows:
\begin{equation}
    H(I)\psi=-h^{2}(\frac{\partial^{2}\psi }{\partial x^{2}} + \frac{\partial^{2}\psi }{\partial y^{2}})-I\psi,
\end{equation}
and $L_{\gamma}^{cl}$ is the suitable semi-classical constant defined as follows:
\begin{equation}
     L_{\gamma}^{cl}=\dfrac{1}{(2\sqrt{\pi})^{2}} \dfrac{\Gamma\left ( \gamma+1 \right )}{\Gamma\left ( \gamma + 2 \right )},
    \label{}
 \end{equation}
\noindent where $\Gamma$ is the Gamma function.  
\end{definition}

 \paragraph*{Remark}
 \noindent As described in previous studies \cite{Laleg-Kirati2013, Kaisserli2014}, parameter $h$ plays a critical role in the overall performance of reconstruction and denoising using SCSA. It was found that when $h$ tends to 0, SCSA uses more negative eigenvalues in the reconstruction, allowing for better convergence to the original image. For a fixed $\gamma$ value, we have 
 
\begin{equation}
   \lim_{h \to 0}  I_{h,\gamma} = I.
   \label{Limit scsa}
\end{equation}
 
 \noindent However, in practice, after discretization, the number of negative eigenvalues is also dependent on the image resolution \cite{evan}. Therefore, an interval for selecting the optimal $h$ in the one-dimensional case was proposed in \cite{evan}, where the upper bound is based on perturbation theory, while the lower bound is based on the sampling theorem. More recently, following the same concept, an optimal method for selecting $h$ for image reconstruction was proposed \cite{Vargas}. However, it should be noted that the selection of an optimal value of $h$ may differ depending on the application, as demonstrated in several studies \cite{Chahid2018,abderrazak2,li}, which reported that the optimal $h$ for signal or image denoising must be defined differently than that reconstruction.

\subsection{Numerical implementation: Windowed pixel-wise 2D-SCSA}

\begin{figure*}
    \centering
    \includegraphics[width=6.2 in,height=2 in]{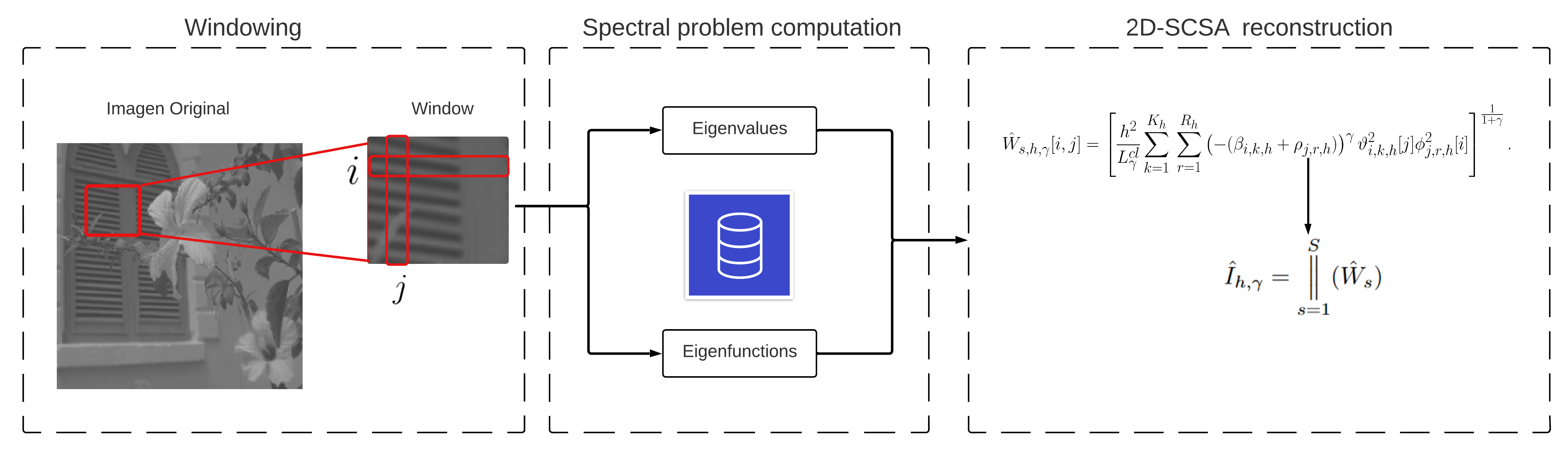}
    \caption{Separation of variables approach for the two-dimensional Semi-Classical Signal Analysis (2D-SCSA) computation. }
    \label{separation_of_variable}
\end{figure*}

\noindent As has been discussed in previous works \cite{Chahid2018,abderrazak2,Vargas}, 2D-SCSA requires the computations of the spectrum of a 2D Schr\"odinger operator, leading to a high computational complexity. For this reason, in \cite{Kaisserli2014} a separation of variables was proposed where the 2D eigenspectrum problem is divided into solving eigenspectrum problems for the rows and columns of the image by using  1D operators. However, this implementation suffers the limitation of only working with square images due to the element-wise matrix operation used to combine the row and column information\cite{Kaisserli2014,Vargas}. For this reason, a new way to compute the 2D-SCSA called  Windowed pixel-wise 2D-SCSA is proposed to solve this limitation (figure \ref{separation_of_variable}), given that the images used tend not to be restricted to only square images. This will allow us to apply the proposed method to any images without losing information caused by any geometrical transformation. The idea is first to separate the image into small windows $\{W_1, W_2, \dots, W_S\}$, with the idea of reducing the computation time raised by solving the eigenvalue problem for big images. Then, for each window ($\hat{W_s}$), the eigenvalues and eigenvectors of each column and row are computed and stored. Finally, the window is reconstructed by applying the 2D-SCSA equation for each pixel.\\

\noindent  Lets define the following 1D Schr\"odinger operators for the window $W_s$:
 
 \begin{equation}
    A_{2,x_0,h}(W_s(x_0,y))=-h^{2}\frac{d^{2}}{d y^{2}}-\frac{1}{2}W_s(x_0,y),
    \end{equation}
    
    \begin{equation}
    B_{2,y_0,h}(W_s(x,y_0))=-h^{2}\frac{d^{2}}{d x^{2}}-\frac{1}{2}W_s(x,y_0),
    \end{equation}
 such that for $(x,y) = (x_0 , y_0)$, the 2D Schrödinger operator $ H(I)$ evaluated at $(x_0 , y_0 )$ is the sum of the operators $A_{2,x_0,h}(W_s(x_0,y))$  and $B_{2,y_0,h}(W_s(x,y_0))$. 
 
 \noindent Subsequently, the spectral problem for these operators is defined as follows:
 \begin{equation}
A_{2,x_0,h}(W_s(x_0,y))\vartheta_{h}^{2}(y_0)=\beta_{x_0,h}\vartheta_{h}^{2}(y_0),
\label{operator A}
\end{equation}
 \begin{equation}
B_{2,y_0,h}(W_s(x,y_0))\phi_{h}^{2}(x_0)=\rho_{y_0,h}\phi_{h}^{2}(x_0),
\label{operator B}
\end{equation}
\noindent where $\beta_{x_0,h}$ and $\vartheta_{h}^{2}(x_0)$ are the negative eigenvalues and $L_2$-normalized eigenfunctions of the operator $A_{2,x_0,h}(W_s(x_0,y))$ and $\rho_{y_0,h}$ and $\phi_{h}^{2}(y_0)$ are the negative eigenvalues and $L_2$-normalized eigenfunctions of the operator $B_{2,y_0,h}(W_s(x,y_0))$ respectively. We use $K_{h}$ (resp. $R_{h}$)  to denote the number of negative eigenvalues for the two previous operators respectively.

\noindent  Then, the 2D representation of the window $W_s$ is estimated at $(x_{0},y_{0})$ using a tensor product of the spectral values  for the 1D  Schr\"odinger operators, as follows:  
 \begin{equation}
 \begin{array}{ll}
     \hat{W}_{i,h,\gamma}(x_0,y_0)=\left[  \dfrac{h^{2}}{L_{\gamma}^{cl}}\mathlarger{\sum}_{k=1}^{K_{h}}\:\mathlarger{\sum}_{r=1}^{R_{h}}\left ( -(\beta_{x_0,k,h}+\rho_{y_0,r,h}) \right )^{\gamma} \vartheta_{x_{0},k,h}^{2}(y_0)\phi _{y_{0},r,h}^{2}(x_0)\right ]^{\frac{1}{1+\gamma}}.
 \end{array}
 \end{equation}

\noindent Schr\"odinger operators (\ref{operator A}) and (\ref{operator B}) are discretized using a Fourier pseudospectral method as follows: 
 \begin{equation}
    -h^{2}D_2\underline{\vartheta}_i-diag\left(\frac{1}{2}W_s[i,:]\right)\underline{\vartheta}_{h,i}=\beta_{h,i}\underline{\vartheta}_{h,i},
    \label{eig_matrix_A}
    \end{equation}
 \begin{equation}
    -h^{2}D_2\underline{\phi}_j-diag\left(\frac{1}{2}W_s[:,j]\right)\underline{\phi}_{h,j}=\rho_{h,j}\underline{\phi}_{h,j},
    \label{eig_matrix_B}
    \end{equation}
\noindent where $D_2$ is a second-order differentiation matrix obtained using the Fourier pseudospectral method, and $diag\left(\frac{1}{2}W_s[i,:]\right)$ and $diag\left(\frac{1}{2}W_s[:,j]\right)$ are the diagonal matrices for the ith row and jth column, respectively. It should be noted that we take $\frac{1}{2}$ of the image potential since each pixel is considered twice in actual implementation. $\underline{\vartheta}=\left[ \vartheta_{1},\vartheta_{2},\dots,\vartheta_{Q-1},\vartheta_{Q}\right]^T$ and $\underline{\phi}=\left[ \phi_{1},\phi_{2},\dots,\phi_{Q-1},\phi_{Q}\right]^T$ are the eigenvectors of the image with size $Q\times Q$.
\noindent The SCSA representation of the window $W_s$ is obtained pixel by pixel using the following formula:
\begin{equation}
 \begin{array}{ll}
     \hat{W}_{i,h,\gamma}[i,j]=\left[  \dfrac{h^{2}}{L_{\gamma}^{cl}}\mathlarger{\sum}_{k=1}^{K_{h}}\:\mathlarger{\sum}_{r=1}^{R_{h}}\left ( -(\beta_{i,k,h}+\rho_{j,r,h}) \right )^{\gamma} \vartheta_{i,k,h}^{2}[j]\phi _{j,r,h}^{2}[i]\right ]^{\frac{1}{1+\gamma}}.
 \end{array}
 \label{2D1DSCSA}
 \end{equation}

\noindent Finally, The reconstructed image $\hat{I}_{h,\gamma}$ is obtain by concatenating all the reconstructed $\hat{W}_n$ windows.

\noindent where $S$ is the number of windows selected to split the image.

\noindent  The 2D-SCSA  algorithm is summarized as follows:
\begin{algorithm}[H]
\caption{Windowed pixel-wise 2D-SCSA}\label{alg:alg1}
\begin{algorithmic}
\STATE 
\STATE {. Initialize} parameters $h$.
\STATE {. Divide the image into windows $\{W_1, W_2, \dots, W_S\}$}.
\STATE \text{. Find} the eigenvalues and eigenvectors for the spectral problems with potentials $\dfrac{1}{2}I[i,:]$ corresponding to the rows and columns from the window $W_s$, using equation \ref{eig_matrix_A} and \ref{eig_matrix_B}.
\STATE  \text{. Calculate} the representation $\hat{W_s}$ for the window $W_s$, using equation (\ref{2D1DSCSA}).

\STATE  \text{. Concatenate} the representations $\{\hat{W_1}, \hat{W_2}, \dots, \hat{W_S}\}$ to generate the image reconstructed  $\hat{I}_{h,\gamma}$.

\end{algorithmic}
\label{alg1}
\end{algorithm}

\subsection{Effect of parameter \texorpdfstring{\(\gamma\)}{gamma} on 2D-SCSA}\label{gamma_effect}

\begin{figure*}[ht]
\centering
\subfloat[]{\includegraphics[width=1.1in]{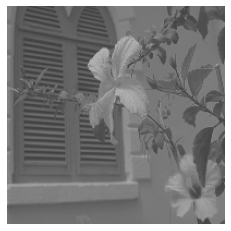}}%
\hspace{0.2 in}
\subfloat[]{\includegraphics[width=1.1in]{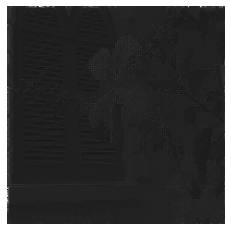}}%
\hspace{0.2 in}
\subfloat[]{\includegraphics[width=1.1in]{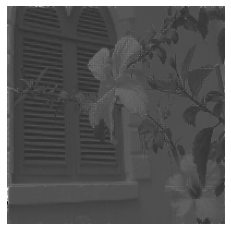}}%
\hspace{0.2 in}
\subfloat[]{\includegraphics[width=1.1in]{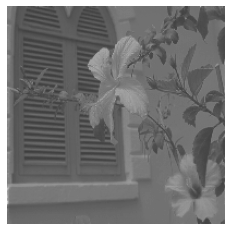}}%
\hspace{0.2 in}
\subfloat[]{\includegraphics[width=1.1in]{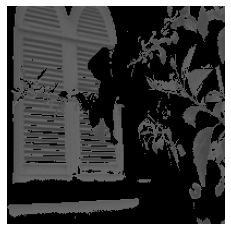}}%

\subfloat[]{\includegraphics[width=1.1in]{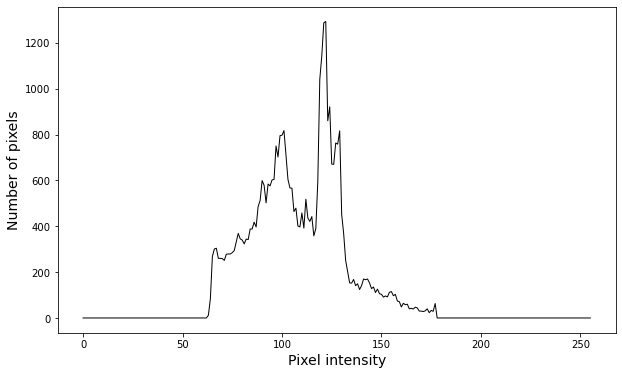}}%
\hspace{0.2 in}
\subfloat[]{\includegraphics[width=1.1in]{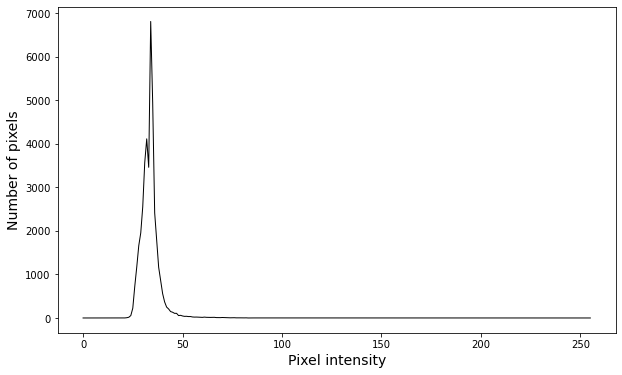}}%
\hspace{0.2 in}
\subfloat[]{\includegraphics[width=1.1in]{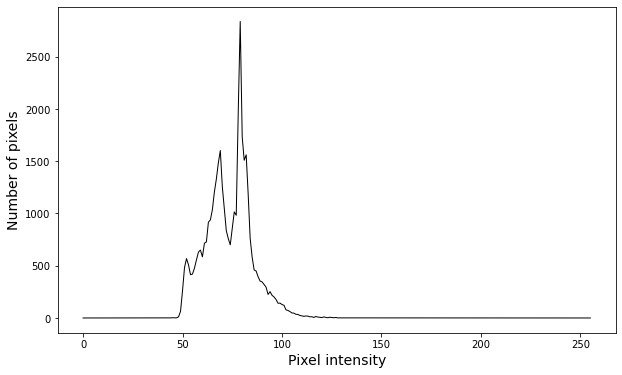}}%
\hspace{0.2 in}
\subfloat[]{\includegraphics[width=1.1in]{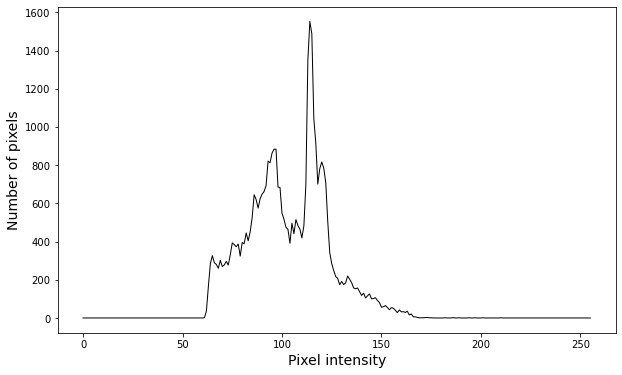}}%
\hspace{0.2 in}
\subfloat[]{\includegraphics[width=1.1in]{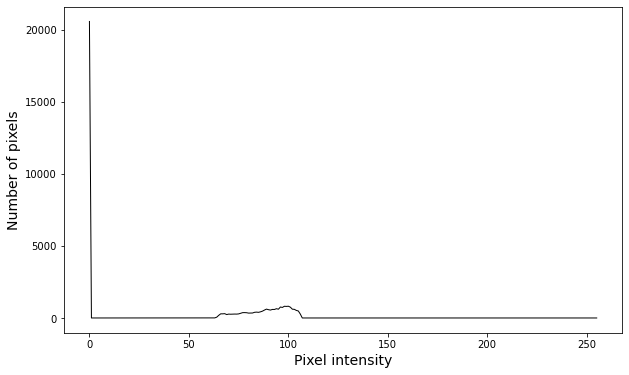}}%

\caption{Effect of the $\gamma$ parameter of the 2D-SCSA for image reconstruction using $h=1$. (a) Original. (b) $\gamma=0.5$.(c) $\gamma=2$.(d) $\gamma=8$.(e) $\gamma=18$. It can be seen that the intensity of the pixels starts to increase with the increase of the parameter $\gamma$, while also the contrast of the pixel increases. Finally, it can be seen that higher values of $\gamma$ caused a loss of the image information by creating artifacts in the images. }
\label{g_effect}
\end{figure*}

\noindent In this section, we analyze the effect of parameter $\gamma$ on 2D-SCSA, which has not been previously studied. Only several tests have been reported for the 1D SCSA case  \cite{HelLal10}, demonstrating that $\gamma = \frac{1}{2}$ is the most appropriate value for 1D signal reconstruction and filtering. However, in \cite{Kaisserli2014}, it was found that higher values of $\gamma$ produces good results for image reconstruction and filter. 

Figure \ref{g_effect} shows the effect of parameter $\gamma$ for image reconstruction. It can be seen that $\gamma$ affects the pixel intensity, changing the global intensity of the image as follows: For high values of $\gamma$, the pixel intensity increases and decreases for lower values of $\gamma$. This is similar to the behavior observed in the gamma transformation method that applies an exponential transformation over the pixels of an image and is described as:
\begin{equation}
    I_{out}=\bar{C}(I_{in})^{\bar{\gamma}},
\label{gamma transfomation}    
\end{equation}

\noindent where $\bar{C}$ is a positive scaling constant commonly set to $1$, and $\bar{\gamma}$ is a positive constant that controls the degree of exponential effect on the pixel intensity. \\

\noindent Similar to the case of the gamma transformation, the $\gamma$ value in 2D-SCSA controls the pixel intensity of an image by applying an exponential effect to reconstructed pixels, as illustrated in equation (\ref{2D1DSCSA}). The main explanation behind this behavior is that $\gamma$ is caused by changes in the amplitude of the negative eigenvalues used to reconstruct the signals produced by the $\gamma$ selected, helping to increase the intensity difference between pixels with similar intensities. However, it's very important to notice that the effect of $\gamma$ for SCSA is more complex since $\gamma$ affects not only the direct eigenvalues but also the whole pixel reconstruction by scaling and applying a similar power effect to the whole reconstruction.\\

\begin{figure}[ht]
\centering
\subfloat[]{\includegraphics[width=1.1in]{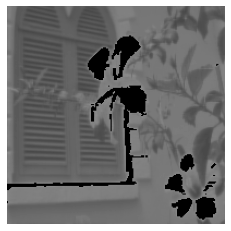}}%
\hspace{0.2 in}
\subfloat[]{\includegraphics[width=1.1in]{images/imgSCSA_toy_g_18.png}}%
\hspace{0.2 in}
\subfloat[]{\includegraphics[width=1.1in]{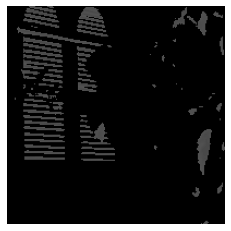}}%
\hspace{0.2 in}
\subfloat[]{\includegraphics[width=1.1in]{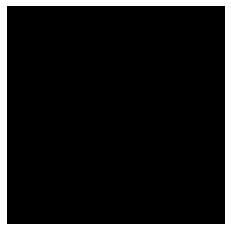}}%

\caption{Effect of the upper limit of the $\gamma$ parameter for the 2D-SCSA for image reconstruction using $h=1$. (a) $\gamma=16$.(b) $\gamma=18$.(c) $\gamma=19$.(d) $\gamma=20$. It can be seen that with the increase of the $\gamma$, the pixels with the higher intensity are converted to NaN until all the pixels are converted to this value.}
\label{g_effectlim}
\end{figure}

\noindent However, a difference between the two methods is that the value of $\gamma$ of 2D-SCSA has an upper limit as it can be seen in figure \ref{g_effectlim}. This upper limit is generated for the $L_{\gamma}^{cl}$. This limit is seen more in the implementation, where the Gamma functions presented in the equation ($\Gamma$($\gamma$+1) and $\Gamma$($\gamma$+2)) will give Inf value given the rapid growth of the function, and the maximum float point precision defined by the IEEE 754 standard for double precision that is around $1.8X10^{308}$. For this reason, when the value of the Gamma function surpasses that limit, the function gives the result Inf and the whole image is full of NaN values.

\begin{figure}[ht]
\centering
\subfloat[]{\includegraphics[width=1.2 in]{images/imgh_toy.png}}%
\subfloat[]{\includegraphics[width=2in]{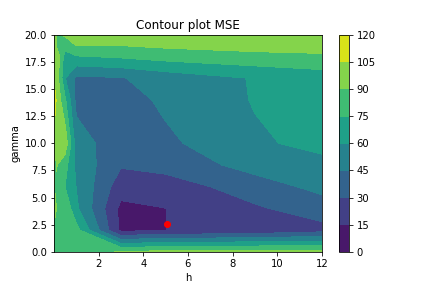}}%
\subfloat[]{\includegraphics[width=2in]{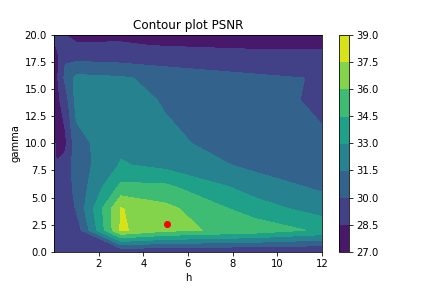}}%
\caption{Relation between $\gamma$ and $h$ values for image reconstruction using 2D-SCSA. (a) Original image. (b) MSE surface plot. (c) PSNR surface plot. This figure shows the relationship between the two parameters of the 2D-SCSA for some well-known reconstruction metrics used to evaluate the performance in the reconstruction and filtering capacity between the original images and the image generated by 2D-SCSA. The red dot indicates the values of $h_{min}$ and $\gamma$ for the image calculated as is proposed in \cite{Vargas}.}
\label{cont_gefect}
\end{figure}

\noindent Finally, it is important to highlight that the pixel intensity of the reconstructed image using 2D-SCSA is affected not only by $\gamma$ but also by $h$, which acts as a scaling factor for the reconstruction. Figure 
\ref{cont_gefect} shows the relation between the $h$ and $\gamma$ parameters for image reconstruction and filtering. As it can be seen, in order to obtain the best performance of the 2D-SCSA, it is necessary to also tune the parameter $\gamma$. Therefore, in this project,  the optimal combination of parameters $\gamma$ and $h$ that help to increase the contrast of the image is selected.

\section{Contrast enhancement using \texorpdfstring{\(\gamma\)}{gamma}-SCSA} \label{contrast scsa}

\noindent In this study, a novel extension of 2D-SCSA called $\gamma$-SCSA is proposed with the main idea of use for the first time this method for contrast enhancement of gray and color images. This extension is based on the parameter $\gamma$ (Section \ref{gamma_effect}), where $\gamma$ values are optimally selected in different regions of the image, leading to enhanced image contrast.

\begin{prop}

\noindent Let $I$ be a low contrast image, and $P$ a prior image generated with the main idea of separate the pixel images into $K$ different groups of pixels ($C_k$), with $k=1,\cdots,K$. 

\begin{equation}
    P=C_{1}\cup C_{2} \cdots \cup C_{K} 
\end{equation}

\noindent These groups are composed of pixels with similar properties such as pixel intensity or texture, allowing for better manipulation of the different intensities present in the image, that will create better and more precise image contrast enhancement. \\

 \noindent Then, the values of $\gamma_k$, $k=1,\cdots, K$ will change for each group $C_k$, as follows:

\[
\gamma_k = \begin{cases} 
      \gamma_1 & \forall \quad W_s[i,j] \in C_1, \\
      \gamma_2 & \forall \quad W_s[i,j] \in C_2, \\
      \vdots\\ 
      \gamma_K & \forall \quad W_s[i,j] \in C_K.
   \end{cases}
\]

\noindent Subsequently, the image $I$ is divides into small windows $W_s$ with $s=1, 2, \dots, S$, and the  contrast-enhanced representation for each window $\hat{W}_{s,h,\gamma_k}$ using the $\gamma$-SCSA is defined as follows:
\begin{equation}
 \begin{array}{ll}
     \hat{W}_{s,h,\gamma_k}[i,j]=\\\left[  \dfrac{h^{2}}{L_{\gamma_k}^{cl}}\mathlarger{\sum}_{u=1}^{U_{h}}\:\mathlarger{\sum}_{r=1}^{R_{h}}\left ( -(\beta_{i,u,h}+\rho_{j,r,h}) \right )^{\gamma_k} \vartheta_{i,u,h}^{2}[j]\phi _{j,r,h}^{2}[i]\right ]^{\frac{1}{1+\gamma_k}}, \\
     k=1,\cdots,K,
 \end{array}
 \label{gSCSA}
 \end{equation}

\noindent where $\gamma_k$ $k=1,\cdots, K$ represents the $\gamma$ value associated with the group $C_k$, $K$ denotes the total number of groups, and $L_{\gamma_k}^{cl}$ is the suitable semi-classical constant defined as follows:
\begin{equation}
     L_{\gamma_k}^{cl}=\dfrac{1}{(2\sqrt{\pi})^{2}} \dfrac{\Gamma\left ( \gamma_k+1 \right )}{\Gamma\left ( \gamma_k + 2 \right )}.
    \label{Laa}
 \end{equation}

\noindent The final image $\widehat{I}_{h,\gamma_k}$ is obtained by concatenating all the $\hat{W}_{s,h,\gamma_k}$ to obtain the image reconstruction $I_{h,\gamma_k}$ and normalizing it as follows:
\begin{equation}
    \widehat{I}_{h,\gamma_k}=\frac{\left [ I_{h,\gamma_k}-min(I_{h,\gamma_k}) \right ]}{\left [max(I_{h,\gamma_k})-min(I_{h,\gamma_k})\right ]}
\end{equation}
\end{prop}

\noindent \paragraph*{Remark}
 This final normalization has been employed in several previous studies \cite{Al,Shi,Jeon,Park,sarkar} to preserve the relative intensity relationship and avoid pixel saturation.\\

\begin{figure*}[ht]
\centering
\subfloat[]{\includegraphics[width=1.6in]{images/imgh_toy.png}%
}
\hfil
\subfloat[]{\includegraphics[width=1.6in]{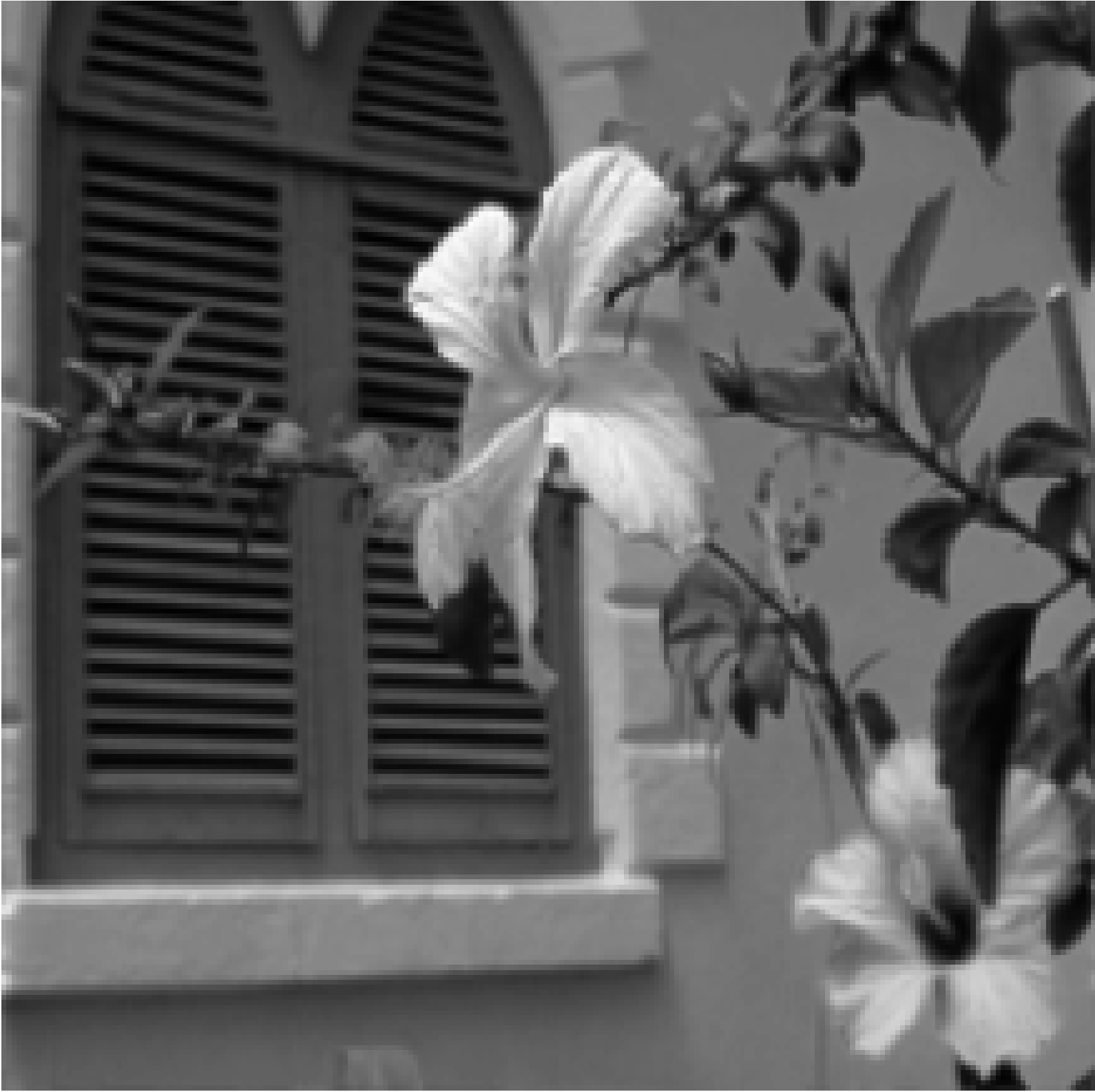}%
}

\subfloat[]{\includegraphics[width=1.6 in]{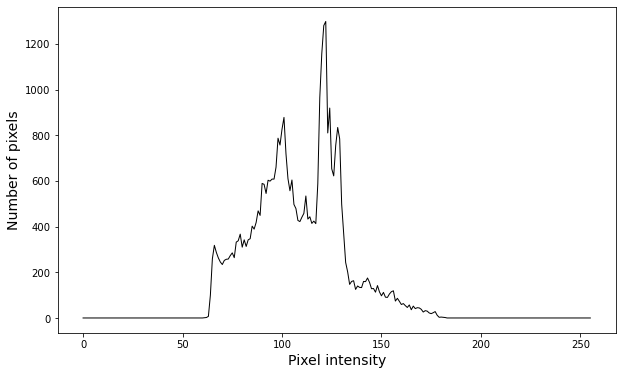}%
}
\hfil
\subfloat[]{\includegraphics[width=1.6in]{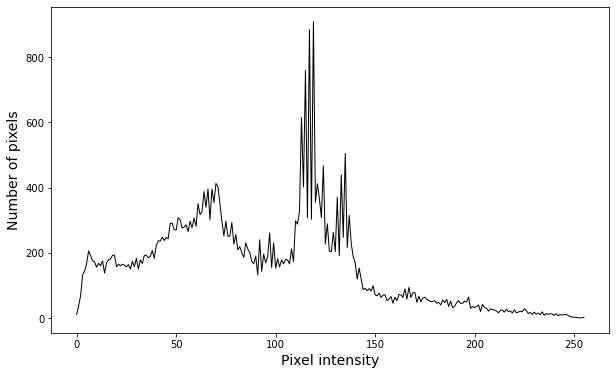}%
}

\caption{ Contrast enhancement using $\gamma$-SCSA. (a)Original image. (b) Enhanced image. (c) Original histogram. (d) Enhance histogram. It can be seen that the $\gamma$-SCSA increases the contrast of the image increasing the difference between the pixels of the different objects in the images improving the quality of the image.}
\label{g_resu}
\end{figure*}

\noindent Based on that, it can be seen that $\gamma$-SCSA exhibits a pseudo-local behavior because different $\gamma$ values are applied for each group of pixels $C_k$ defined by the prior, enabling local and independent manipulation of the pixel intensities of each group (Fig. \ref{g_resu}). Meanwhile, parameter $h$ continues to be applied globally to all pixels of the image. The main concept behind $\gamma$-SCSA is to increase the image contrast by manipulating the pixel intensity of each group $C_k$ independently from that of other groups. \\

\begin{figure*}[t]
\centering
\includegraphics[width=6.2 in,height=2.5 in]{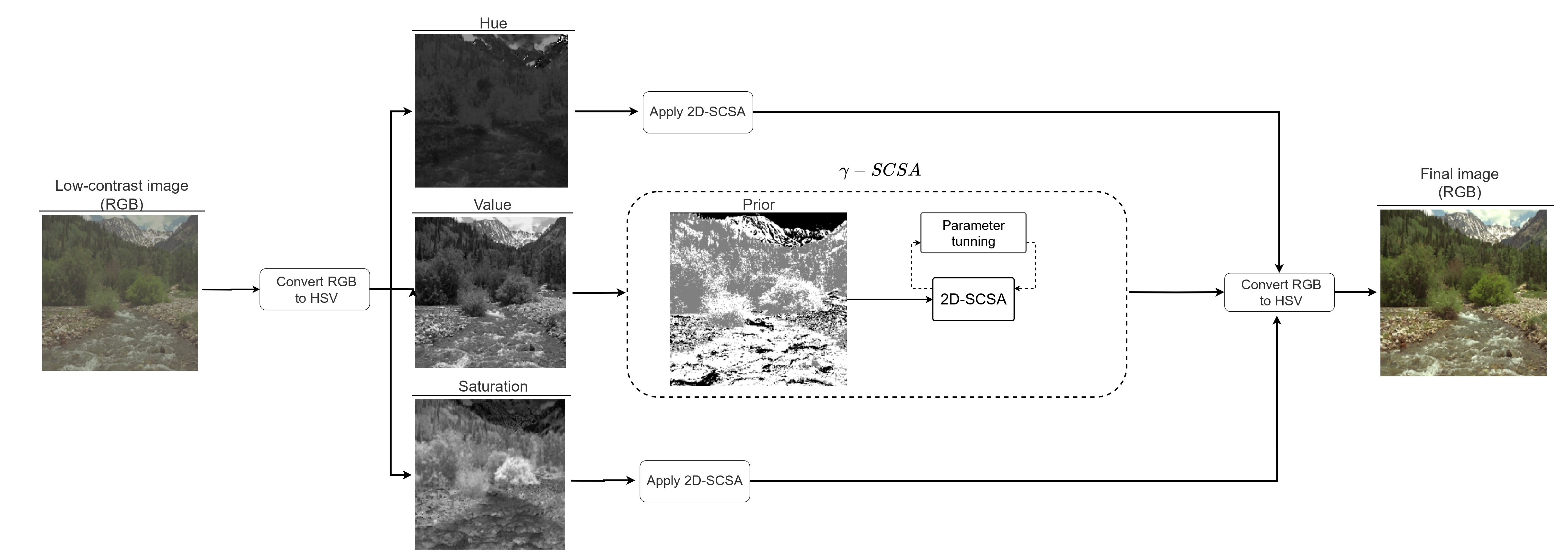} 
\caption{Schematic illustration of the contrast enhancement method based on the $\gamma$-SCSA.}
\label{fram}
\end{figure*}

\section{Contrast enhancement algorithm}

\noindent In this study, the proposed $\gamma$-SCSA method is used for contrast enhancement of color images, in contrast with previous papers where the 2D-SCSA was applied only to grayscale images \cite{Kaisserli2014,Vargas,Chahid2018} proven the good performance of 2D-SCSA for image reconstruction, filtering and features extraction. The procedure of the proposed method is illustrated in Fig. \ref{fram}, where the image is first decomposed into the hue–saturation–value (HSV) color space allowing the independent manipulation of the brightness while reducing the probability of color distortion \cite{Arici,Kuran}. In particular, the Value (V) channel was used because ensures that the color information of the image remains unchanged while the contrast is enhanced. After this transformation, the prior image $P$ is computed based on the V channel using the three different methods proposed in this study. Subsequently, $\gamma$-SCSA is applied to this channel to increase the contrast. In addition, the 2D-SCSA is applied for the channel $H$ and $S$ to filter noise that may be presented in the original image. Finally, the image is converted back to red–green–blue color space.

\subsection{\texorpdfstring{\(\gamma\)}{gamma} prior} \label{ggroups}
\noindent  The selection of different groups of pixels from the original image is a critical step in the proposed contrast enhancement method since allows for control of the pixel intensity belonging to different parts of the images by applying several values of $\gamma$ during the reconstruction. In addition, the capacity to control the intensity, helps to decrease the possibility to present over-enhancement of pixels with a high initial intensity and allows to obtain a more local enhancement that improves the quality of the final result. In this study, different algorithms were tested to create the prior grouping of $\gamma$, intending to select the grouping method that presents a better capacity to handle low-contrast noisy images used in this study. The algorithms used in this project are:\\

\subsubsection{Weighted Multitask  Fuzzy C-means (WMT-FCM)}

Fuzzy C-Means (FCM) is a clustering algorithm extensively used in medical image processing for tasks such as image segmentation \cite{c-fuzzy1,fuzzy-fet}. Unlike hard clustering techniques such as k-means, FCM allows for partial membership in multiple clusters. This flexibility is particularly advantageous in low-contrast images, where boundaries between different objects in the image can be ambiguous. 
Taking into account the advantages of the fuzzy C-means method, in this study, the clustering algorithm proposed in 2023 by \cite{c-fuzzy1} is used as a prior for the the $\gamma$ distribution. This method integrates fuzzy C-means with multitask learning strategy, leading to the development of the Weighted Multitask Fuzzy C-means (WMT-FCM) algorithm. This algorithm is based on the capacity to share information across multiple related segmentation tasks while also capturing specific information related to each task. This mechanism dynamically assigns optimal weights to each task, ensuring that tasks with better clustering effects contribute more significantly to the shared public information. This allows the algorithm to maintain high performance even in varying noise levels and complex intensity variations across the images.

\subsubsection{Frequency-Tuned Gaussian Mixture Model (FTGMM)}
Gaussian Mixture Models (GMM) is a sophisticated approach for low contrast image segmentation. GMM assumes that the data is a mixture of several Gaussian distributions, each representing a cluster. This model is particularly useful in cases where the different objects in the image present similar intensities. By fitting a GMM to the pixel intensity values, it is possible to determine the probability of belonging of each pixel to a cluster, allowing for fine-grained contrast adjustment as well as to understand.

\noindent In this study, The FTGMM (Frequency-Tuned Gaussian Mixture Model) method proposed in \cite{Pan} is used as a prior for $\gamma$-SCSA. The main idea of this method is to incorporate frequency-tuned salient region detection into the Gaussian Mixture Model (GMM). As a result, this gives a saliency map that highlight the most important parts of an image. This saliency map weighs the pixels, emphasizing important areas during segmentation. The method effectively segments the image into distinct regions by iteratively adjusting the GMM parameters (means, covariances, and weights) based on the saliency-weighted pixel values. The main advantage of this approach is that it combines the spatial information from the saliency map and the statistical modeling of GMM to achieve precise, and robust segmentation in complex images.\\

\noindent Finally, a number of 4 clusters for each method were selected based on a sensitivity analysis made using the Davies-Bouldin Index to select the best number of clusters. Then, each of the methods will be used as prior for the $\gamma$ distribution of the 2D-SCSA, giving spatial information about how pixels are distributed, allowing a better contrast enhancement at the same time that provides a way to interpret the result obtained for the 2D-SCSA. This is important because that will allow clinicians and researchers to understand how the proposed algorithm distributes the $\gamma$ across the image, as well as, help to understand possible limitations of the method.\\

\subsubsection{Prior selection}

\begin{figure*}[ht]
\centering
\subfloat[]{\includegraphics[width=1.5in]{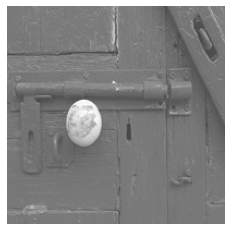}%
}
\hfil
\subfloat[]{\includegraphics[width=1.5in]{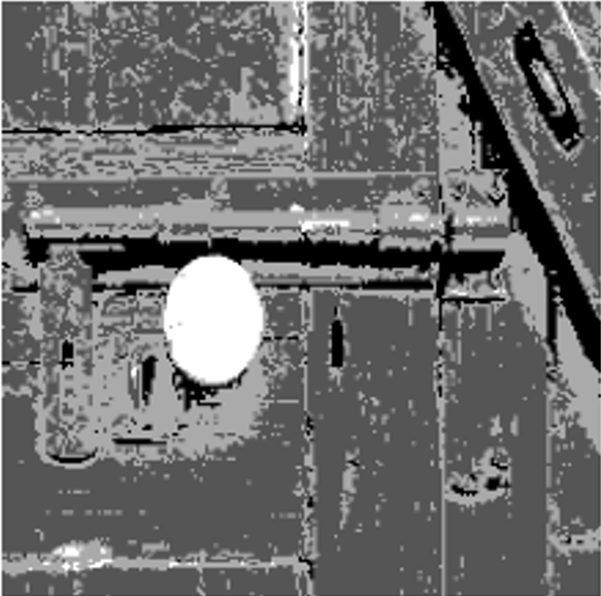}%
}
\hfil
\subfloat[]{\includegraphics[width=1.5in]{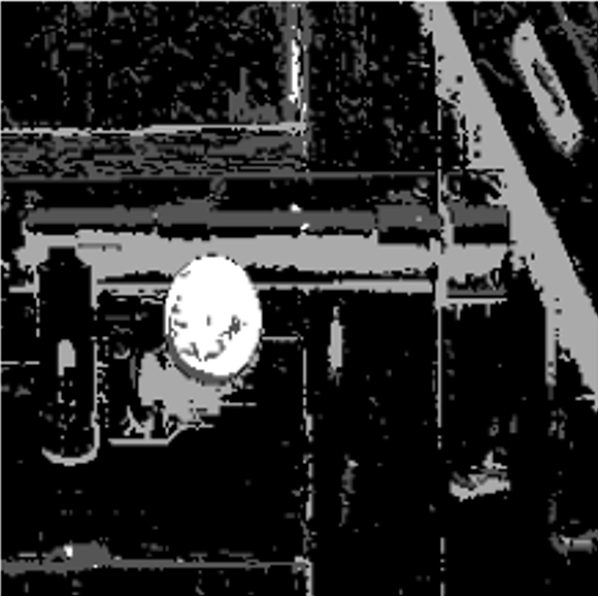}%
}

\subfloat[]{\includegraphics[width=1.5 in]{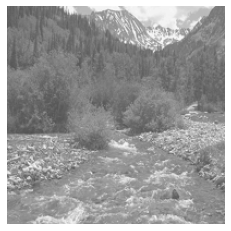}%
}
\hfil
\subfloat[]{\includegraphics[width=1.5in]{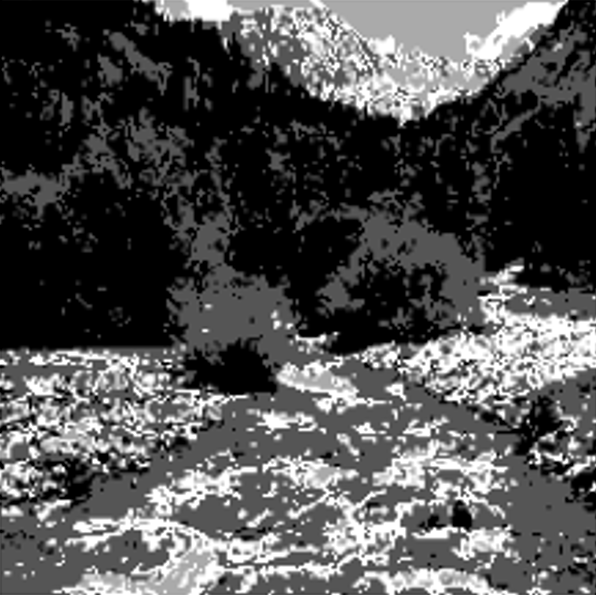}%
}
\hfil
\subfloat[]{\includegraphics[width=1.5 in]{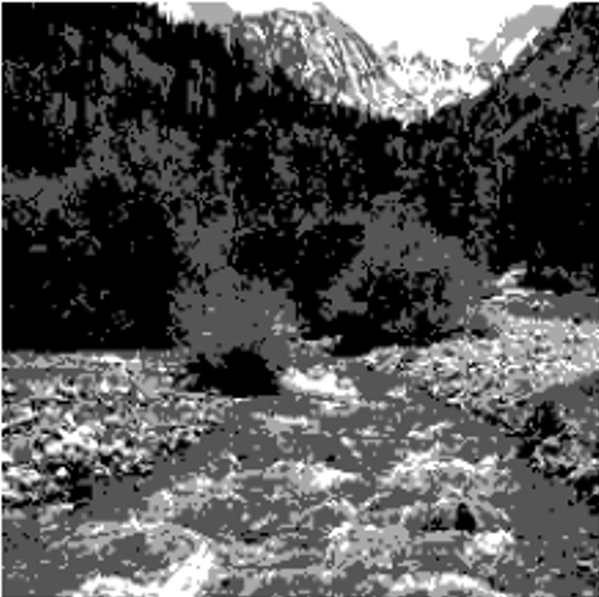}%
}

\caption{Prior image ($P$) . (a)Original image. (b) Image obtained using FTGMM. (c) Image obtained using WMT-FCM. The Prior images are obtained based on the Value channel of the HSV color channel using 4 clusters for each method.}
\label{prior}
\end{figure*}

As it was mention in previous sections, the prior selection plays a crucial role in the performance of the proposed method. Figure \ref{prior} shows the results of the different algorithms used on the Value channel. As can be seen, the WMT-FCM is capable of maintaining more fine details compared to the FT-GMM method. This can be seen in some details such as the difference in the door handle where the WMT-FCM is capable of showing some of the subtle stains presented, while, the FT-GMM its not capable of capture it. This same behavior can be seen in images f and e where the WMT-FCM presented a better capacity to represent the structure of the pixels of the mountain. This is very important because will help to improve the results of the $\gamma$-SCSA allowing us to maintain the general structure of the image as well as reducing the possibility of present artifacts.

\noindent In addition to this qualitative analysis, a quantitative analysis was made by computing the mean Davies-Bouldin Index for each of the datasets proposed. This index measure how well the clusters are separated between them and how compact is each of the clusters, where lower values indicates a better performance of the clustering method.

\begin{table}[!htp]\centering
\caption{mean Davies-Bouldin Index for TID13 and CSQI datasets}\label{tab:prior}
\scriptsize
\begin{tabular}{lrrr}\toprule
Dataset &WMT-FCM &FT-GMM \\\midrule
TID13 &0.46 &0.52 \\
CSQI &0.41 &0.44 \\
\bottomrule
\end{tabular}

\end{table}

\noindent At its can be observed at Table \ref{tab:prior}, the difference between the methods its not very big, however, WMT-FCM obtain a better perform in both datasets. This performance can be explained by the fuzzy nature of the WTM-FCM that allow this method to handle better the uncertainty given by the low contrast between the different pixels. For this reason, the WTM-FCM method was selected as prior for the $\gamma$-SCSA.

\subsection{Multi-objective optimization}\label{GA}
\noindent As discussed in previous studies \cite{Laleg-Kirati2013,Kaisserli2014}, the performance of the SCSA is highly dependent on the choice of parameter $h$, with a poor selection leading to poor performance. Therefore, a method to define $h$ for 2D-SCSA has been proposed in a previous study \cite{Vargas} based on the lower limit of $h$ previously proposed for the 1D-SCSA case  \cite{evan}. However, this method was proposed for image reconstruction, where the value of $h$ guarantees a good performance to reconstruct the original image. Nevertheless, in the current study, the main idea is not to reconstruct the original low-contrast image, instead, we want to find the best $h$ value that helps to obtain the high-contrast image. In addition, due to the interaction between $h$ and $\gamma$ explained before, the selection of one parameter will be affected by the value of the other. Therefore, a multi-objective optimization is used to determine the combination of these parameters that produces the optimal contrast enhancement. This approach provides a powerful framework for performing contrast enhancement tasks because it balances enhancing image contrast and preserving important image details \cite{Kuran}, which can be conflicting objectives \cite{YANG}. In this study, the non-dominated sorting genetic algorithm II (NSAG22) was used for multi-objective optimization, making it possible to find a solution by simultaneously considering multiple objectives in the optimization process.\\

\noindent The main principle of NSGA2 is to employ genetic operators, such as crossover and mutation, to generate a diverse set of candidate solutions. These solutions represent different trade-offs between contrast enhancement and detail preservation. Through this process, NSAG2 explores the space of possible solutions, separating the solutions into different fronts based on their dominance. Then, the solutions inside the Pareto front represent the best trade-offs between contrast enhancement and detail preservation. This allows users to choose a solution that best suits their requirements and preferences \cite{Verma}.\\

\subsubsection{Parameter initialization}
\noindent In this study, the initial population was defined randomly within the range defined for each chromosome where the values for the $h$ parameter are defined by $h_{min}$ and the upper limit is $10*h_{min}$. This range has been shown to provide a good searching space for $h$ in previous studies \cite{Vargas,li} and for the value of $\gamma$ the lower limit was set to $0.001$ due to the theory of SCSA where this value needs to be greater than $0$. For the upper value, $35$ was empirically chosen (Table \ref{tab1}).

\begin{table}[H]
\begin{center}
\caption{Range for the initial chromosome values}
\label{tab1}
\begin{tabular}{| c | c | c |}
\hline
Parameter & Min & Max\\

\hline
$h$&  $h_{min}$& $10*h_{min}$\\
\hline
$\gamma$ &0.001 & 35\\ 
\hline

\end{tabular}
\end{center}
\end{table}

\subsubsection{Cost function}

\noindent For the cost function, two objective functions, $J_1$ and $J_2$, were proposed to evaluate the fidelity of reconstructed images and the amount of detail in the image.

\begin{align}
    J_1 &= \alpha \left( 1 - \frac{SSIM(I, \widehat{I}_{h,\gamma_k}) + 1}{2} \right) \nonumber \\
    &+ (1 - \alpha) \left( 1 - \frac{PSNR(\widehat{I}_{h,\gamma_k})}{max\_PSNR} \right)
\end{align}
 \vspace{0.05cm}
\[
    J_2=E(\widehat{I}_{h,\gamma_k}),
\]

\noindent where $I$ represents the original image, and $\widehat{I}_{h,\gamma_k}$
represents the image enhanced by the $\gamma$-SCSA.\\

\noindent The cost function $J_1$ was defined as the combination of two well-known metrics, the Structural Similarity Index (SSIM) and the Peak Signal-to-Noise Ratio (PSNR). The main reason for the combination of these two metrics was to control two important aspects of the SCSA, such as the image reconstruction controlled by the SSIM and the filtering controlled by the PSNR, since one of the main issues of contrast enhancement algorithms is the enhancement of noise. To use the strengths of both SSIM and PSNR, a combined cost function ($J_1$) was created. It is important to mention that given the different scales of both metrics, a normalizing step was introduced for each metric to obtain a common scale and combine the two metrics using a weighted sum controlled by parameter $\alpha$ between $0$ to $1$, that balances the contribution of each parameter. This cost function helped in preventing image distortion, such as noise and artifacts. It helps to achieve a balance between enhancing contrast and minimizing the introduction of unwanted distortions \cite{Luis,VIJA}, resulting in enhanced image quality while maintaining overall visual similarity to the original image.\\

\noindent On the other hand, the cost function ($J_2$) is composed of the entropy ($E$) of the image, which measures the information present on the image. In contrast enhancement, entropy can be used as a criterion to evaluate the enhancement, because higher entropy values indicate that the image possesses more details and has better quality.\\

\noindent The entropy of an image can be calculated using the following equation:

\[ E(\widehat{I}_{h,\gamma_k})= - \sum_{i=0}^{L-1} p(i) \log_2(p(i)), \]

\noindent where \( E \) represents the entropy of image \(\widehat{I}_{h,\gamma_k} \), \( p(i) \) is the probability of occurrence of pixel intensity \( i \), and \( L \) is the total number of possible pixel intensity levels.\\

\noindent  The main idea of using entropy as a cost function was to increase the pixel distribution of the low-contrast image across the dynamic range, helping to balance the representation of dark and bright regions. Then, the algorithm can effectively enhance subtle textures, edges, and fine structures, leading to improved image interpretability \cite{Kuran}.\\

\begin{figure*}[ht]
\centering
\subfloat[]{\includegraphics[width=1.5in]{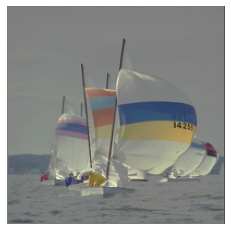}%
}
\hfil
\subfloat[]{\includegraphics[width=1.5in]{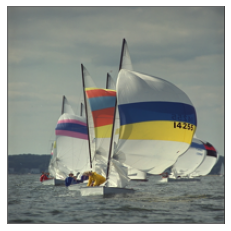}%
}
\hfil
\subfloat[]{\includegraphics[width=1.5in]{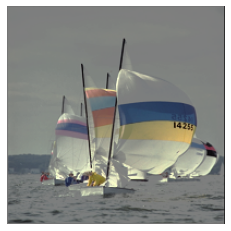}%
}

\subfloat[]{\includegraphics[width=1.5in]{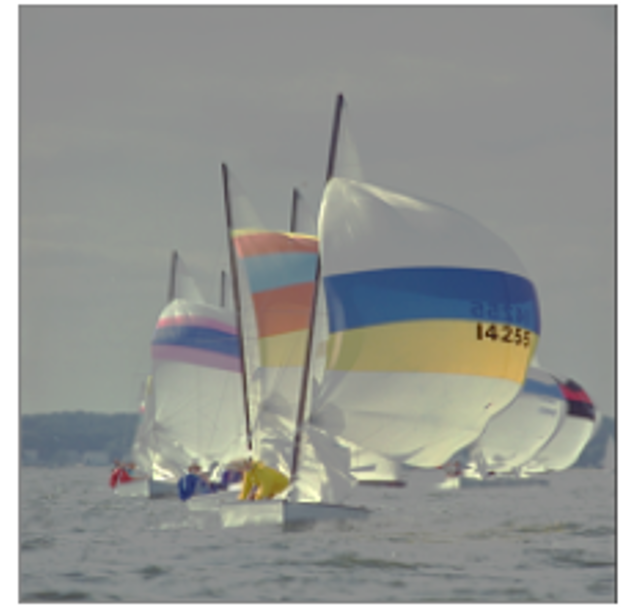}}%
\hfil
\subfloat[]{\includegraphics[width=1.5in]{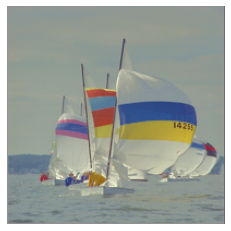}%
}
\hfil
\subfloat[]{\includegraphics[width=1.5 in]{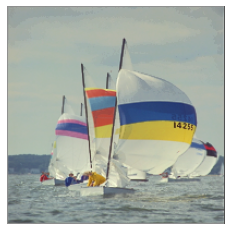}%
}

\caption{Noise-free image contrast enhancement. (a)Original image. (b) ground-truth. (c) IAGCWD (d) THSI
(e) PWGC (d) Ours.}
\label{nf_img}
\end{figure*}

\subsubsection{Parameter selection}

\noindent Finally, for each generation, the two best chromosomes were selected, these chromosomes were taken into a simulated binary crossover stage with a probability of $0.2$, causing similar chromosomes to be generated in the children. Subsequently, a polynomial bounded mutation with a probability of  $0.5$ was used to obtain diverse solutions, to obtain less similar children compared to their parents. Finally, the stopping criteria of the algorithm were set to 10 generations.\\

\noindent A unique solution from the Pareto front was defined using the augmented scalarization function. The main principle of this function is to obtain a single scalar function by aggregating different objective functions using predefined weights \cite{ABOU}. This method enhanced the decision-making process by providing a comprehensive and flexible approach to selecting one unique solution from the Pareto front, allowing us to make a choice that aligns with our objectives and preferences.

\subsection{Evaluation metrics}

\noindent Various metrics have been used in previous studies have been used for the evaluation of contrast enhancement algorithms based on the achievement of adequate contrast enhancement with minimum distortion. To evaluate the performance of the algorithms, the following metrics were applied: Mean Square Error (MSE)\cite{Asamoah2018,
attar}, Peak Signal to Noise Ratio (PSNR)\cite{Asamoah2018}, Structural Similarity Index (SSIM) \cite{MORENO},  Entropy ratio defined as the entropy of the image enhanced over the entropy of the original image \cite{sahnon2018}, Contrast Improvement Index (CII) \cite{Mirza}, Entropy Enhancement Measure (EME)\cite{Zhou2019}, Universal Quality Image Index (UQI)\cite{Singh2018}. Finally, it’s important to
clarify that to have a fair comparison between the learning-based and image-processing methods, all the methods were tested and compared using the same images, corresponding to 30\% of the datasets. In addition, learning solutions where pre-trained with LOL dataset to avoid overfitting due to not enough training data.

\section{Results and discussion}
\noindent The proposed contrast enhancement method was evaluated using two well-established databases used for contrast enhancement evaluation: TID2013, and the CSIQ datasets used for test contrast enhancement methods, where we had an original low contrast image, and the high contrast ground truth that will help us to investigate the performance of the method \cite{Vijaya,Kumar,Pallavi,Kuran,Srinivas}. These databases offer diverse image collections with varying degrees of contrast, enabling comparison with existing techniques and providing insights into the performance of the proposed method. In addition, the results will be compared with some state of the art method such as Tripartite sub-image histogram equalization (TSIHE) \cite{tripa}, Improved adaptive gamma correction with weight distribution (IAGCWD) \cite{IAGCWD}, Pixel-wise gamma correction (PWGC) \cite{PWGC}.

\subsection{Qualitative analysis}

\noindent Figure \ref{nf_img} presents examples obtained by applying the different methods of comparison as well as the proposed $\gamma$-SCSA in noise-free images. It can be seen that the PWGC and the $\gamma$-SCSA, showed the best performance generating the higher quality images. However, it can be appreciated that $\gamma$-SCSA where able to capture small details of the waves in the sea and in the sky. This capacity to capture the small structures can be related to the prior selection that provides spatial information to the SCSA and the capacity of the SCSA as a reconstruction method that will try to reduce the amount of structure loss while enhancing the contrast of the image.

\begin{figure*}[ht]
\centering
\subfloat[]{\includegraphics[width=1.5in]{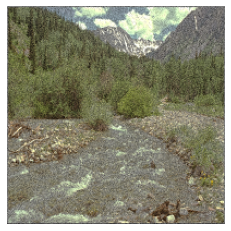}%
}
\hfil
\subfloat[]{\includegraphics[width=1.5in]{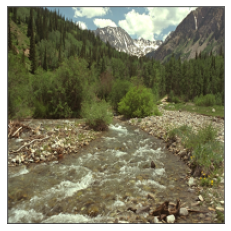}%
}
\hfil
\subfloat[]{\includegraphics[width=1.5in]{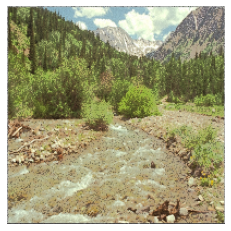}}%

\subfloat[]{\includegraphics[width=1.5in]{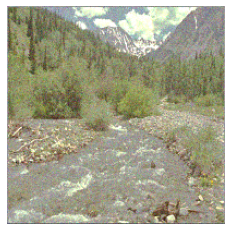}}%
\hfil
\subfloat[]{\includegraphics[width=1.5in]{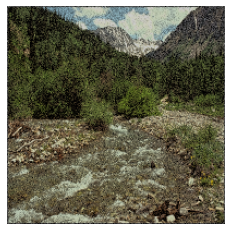}}%
\hfil
\subfloat[]{\includegraphics[width=1.5 in]{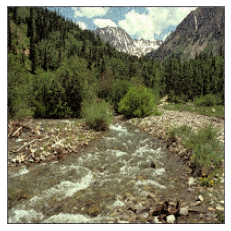}%
}

\caption{Noisy image contrast enhancement (SNR=20dB) . (a)Original image. (b) ground-truth. (c) IAGCWD (d) THSI
(e) PWGC (d) Ours.}
\label{noisy}
\end{figure*}

\noindent In addition, Fig. \ref{noisy} presents the results obtained by applying the methods on noisy images that were obtained by additive white Gaussian noise of 20 dB. It can be seen that THSIE and the IAGCDW methods were the methods that had the worst performance for the contrast of the noise image, producing an image with several artifacts reducing the interpretability of the image. In contrast, the PWGC and $\gamma$-SCSA were able to maintain the different color tones of the images, in comparison the the others methods, where the colors where shifted.\\

\noindent Furthermore, the $\gamma$-SCSA obtains the best image, reducing the presence of noise compared with the other methods, showing the advantage of the proposed method against noisy images. This capacity of SCSA to handle noisy data has been discussed in previous works \cite{abderrazak2,Vargas,Kaisserli2014}, where it was found that by tuning the value of $h$ used to reconstruct the image, this method is capable of filtering high-frequency noise present in the image. For this reason, the proposed $\gamma$-SCSA uses this capacity to handle noise in the image while at the same time enhance the contrast of the image using the effect of $\gamma$. This allows good performance to handle low-contrast noisy images.   

\subsection{Quantitative analsysis}

Table \ref{tab:simu} shows the results obtained for each of the methods proposed for noise-free images. Similar to the observed in Figure \ref{nf_img}, that $\gamma$-SCSA and PWGC methods show a great overall performance. Additionally, It's important to highlight that $\gamma$-SCSA presented the highest SSIM of 0.90 and the lowest MSE of 8.96, demonstrating superior structural similarity and minimal error compared with the ground truth. This shows the effectiveness of preserving the subtle details of the images, helping to minimize possible distortion in color and structure that may occur while working with low-contrast images.Furthermore, Its high entropy (3.55) and UQI (0.90) scores further emphasize its capability to enhance contrast effectively while maintaining image integrity. Similarly, PWGC performed well with similar values for all the metrics studied, this shows the capacity of this learning-based method to capture general intrinsic information present in the image. However, maybe one of the limitations this method could present was the limited amount of data presented. This limitation was diminished by using pre-trained weights obtained by train the model in the LOL dataset, that is a dataset used for low-light image enhancement. These results show the capacity of the proposed method to obtain comparable results with the state of the art and show greater capacity to deal with noisy data. Another advantage of the proposed method is that there is no need to train a machine learning model, which is very useful in scenarios where there is not too much data available. 

\begin{table}[!htp]\centering
\caption{Results obtained for noise-free contrast enhancement }\label{tab:simu}
\small

\begin{tabular}{cccccccc}\toprule
Methods &SSIM &PSNR &MSE &Entropy\_ratio &CII &UQI \\\midrule
PWGC \cite{PWGC} &0.88&28.01 &10.22 &\textbf{3.25} &\textbf{3.95} &0.88 \\
IAGCWD \cite{IAGCWD}&0.70 &20.64 &28.43 &2.96 &2.73 &0.78 \\
THSIE \cite{tripa} & 0.73&22.30 &38.54 &2.77 &2.89 &0.80 \\
$\gamma$-SCSA &\textbf{0.90} &\textbf{32.17} &\textbf{8.96} &3.09 &\textbf{3.52} &\textbf{0.90} \\
\bottomrule
\end{tabular}
\end{table}

\noindent Secondly, the performance of the method was analyzed for the noisy images. As table \ref{tab:noi} shows, the IAGCWD method shows the best performance for the Entropy ratio and the CII metrics, however, for the other metrics the performance wasn't very well. This can occur due to over-enhancement of the image, giving as result an image with too much change of color that provokes a distortion in the image. For this reason, even if the method increase the CII and the entropy in the image, fail to conserve the important details of the images, given as result a non-realistic image. In addition, the PWGC method shows a good performance to maintain the important details of the image, generating a more realistic image. However, the main limitation of this method is the presence of noise in the final image. This can be generated by the fact that the neural network was only pre-trained with low-light images that will not present this class of noise, making it harder for the model the adaptation to the noisy images. For this reason, further optimization or more specialized training could allow to use of the full potential of this architecture for non-reference contrast enhancement.

\begin{table}[!htp]\centering
\caption{Results obtained for noise-free contrast enhancement }\label{tab:noi}
\small

\begin{tabular}{cccccccc}\toprule
Methods &SSIM &PSNR &MSE &Entropy\_ratio &CII &UQI \\\midrule
PWGC \cite{PWGC} &0.79&9.01 &55.22 &3.25 &3.95 &0.80 \\
IAGCWD \cite{IAGCWD}&0.60 &7.43 &74.43 &\textbf{7.96}&\textbf{5.73} &0.68 \\
THSIE \cite{tripa} & 0.54&5.12 &68.54 &3.87 &1.42 &0.49 \\
$\gamma$-SCSA&\textbf{0.83} &\textbf{18.43} &\textbf{44.62} &4.69 &4.02 &\textbf{0.85} \\
\bottomrule
\end{tabular}
\end{table}

\noindent In contrast, $\gamma$-SCSA shows solid performance in all the metrics studied, showing the great capacity of the Weighted Multitask Fuzzy C-means clustering approach as a prior for the $\gamma$ distribution. This method achieves the highest PSNR value of 18.43, showing the strength of the $\gamma$-SCSA method to handle noisy data and at the same time preserve the quality of the image. These qualities are crucial in contrast image enhancement where in some applications is necessary to deal with noisy images, and if the method is not able to adapt to this noise, the result will have artifacts that will affect the final result.

\section{Conclusion}
\noindent In this study, a novel SCSA-based image contrast enhancement method called $\gamma$-SCSA was introduced. $\gamma$-SCSA enhances the contrast of images by applying different $\gamma$ values for the images. The proposed implementation method uses a prior based on WMT-FCM clustering algorithm for selecting different groups of $\gamma$, providing spatial information for the selection of $\gamma$ values and decreasing the probability of present artifacts such as halos, over-enhancement, increased noise, and level saturation  \cite{Srinivas}. The proposed method was evaluated on two of the most commonly used datasets for contrast enhancement, CQSI and TID2013, and achieved good performance. The results demonstrate that $\gamma$-SCSA can enhance the contrast of images while preserving their natural characteristics, producing the desired enhancement with almost no artifacts. The results further indicate the robustness of the proposed method on various standard color images with different levels of contrast degradation. Future work includes adapting the value of parameter h to the noise level of the image and applying the proposed method to low-light image enhancement and medical image contrast enhancement.

\bibliographystyle{unsrtnat}

\begin{thebibliography}{42}
\providecommand{\natexlab}[1]{#1}
\providecommand{\url}[1]{\texttt{#1}}
\expandafter\ifx\csname urlstyle\endcsname\relax
  \providecommand{\doi}[1]{doi: #1}\else
  \providecommand{\doi}{doi: \begingroup \urlstyle{rm}\Url}\fi

\bibitem[Ariateja et~al.(2018)Ariateja, Ardiyanto, and Soesanti]{Ari}
Dananjaya Ariateja, Igi Ardiyanto, and Indah Soesanti.
\newblock A review of contrast enhancement techniques in digital image
  processing.
\newblock \emph{2018 4th International Conference on Science and Technology
  (ICST)}, pages 1--6, 2018.
\newblock \doi{10.1109/ICSTC.2018.8528579}.
\newblock URL \url{https://ieeexplore.ieee.org/document/8528579}.

\bibitem[Srinivas et~al.(2021)Srinivas, Bhandari, and Kumar]{Srinivas}
Kankanala Srinivas, Ashish~Kumar Bhandari, and Puli~Kishore Kumar.
\newblock A context-based image contrast enhancement using energy equalization
  with clipping limit.
\newblock \emph{IEEE Transactions on Image Processing}, 30:\penalty0
  5391--5401, 2021.
\newblock \doi{10.1109/TIP.2021.3083448}.
\newblock URL \url{https://ieeexplore.ieee.org/abstract/document/9444637}.

\bibitem[Zhao et~al.(2023)Zhao, Wang, Guo, and Zhang]{ZHAO}
Runxing Zhao, Zhiwen Wang, Wuyuan Guo, and Canlong Zhang.
\newblock Multi-scene image enhancement based on multi-channel illumination
  estimation.
\newblock \emph{Expert Systems with Applications}, 226:\penalty0 120271, 2023.
\newblock ISSN 0957-4174.
\newblock \doi{https://doi.org/10.1016/j.eswa.2023.120271}.
\newblock URL
  \url{https://www.sciencedirect.com/science/article/pii/S095741742300773X}.

\bibitem[Saleem et~al.(2017)Saleem, Beghdadi, and Boashash]{SALEEM2017161}
Amina Saleem, Azeddine Beghdadi, and Boualem Boashash.
\newblock A distortion-free contrast enhancement technique based on a
  perceptual fusion scheme.
\newblock \emph{Neurocomputing}, 226:\penalty0 161--167, 2017.
\newblock ISSN 0925-2312.
\newblock \doi{https://doi.org/10.1016/j.neucom.2016.11.044}.
\newblock URL
  \url{https://www.sciencedirect.com/science/article/pii/S092523121631459X}.

\bibitem[Pizer et~al.(1987)Pizer, Amburn, Austin, Cromartie, Geselowitz, Greer,
  {ter Haar Romeny}, Zimmerman, and Zuiderveld]{PIZER1987355}
Stephen~M. Pizer, E.~Philip Amburn, John~D. Austin, Robert Cromartie, Ari
  Geselowitz, Trey Greer, Bart {ter Haar Romeny}, John~B. Zimmerman, and Karel
  Zuiderveld.
\newblock Adaptive histogram equalization and its variations.
\newblock \emph{Computer Vision, Graphics, and Image Processing}, 39\penalty0
  (3):\penalty0 355--368, 1987.
\newblock ISSN 0734-189X.
\newblock \doi{https://doi.org/10.1016/S0734-189X(87)80186-X}.
\newblock URL
  \url{https://www.sciencedirect.com/science/article/pii/S0734189X8780186X}.

\bibitem[Kim(1997)]{kim}
Yeong-Taeg Kim.
\newblock Contrast enhancement using brightness preserving bi-histogram
  equalization.
\newblock \emph{IEEE Transactions on Consumer Electronics}, 43\penalty0
  (1):\penalty0 1--8, 1997.
\newblock \doi{10.1109/30.580378}.

\bibitem[Ibrahim and Pik~Kong(2007)]{Ibrahim}
Haidi Ibrahim and Nicholas~Sia Pik~Kong.
\newblock Brightness preserving dynamic histogram equalization for image
  contrast enhancement.
\newblock \emph{IEEE Transactions on Consumer Electronics}, 53\penalty0
  (4):\penalty0 1752--1758, 2007.
\newblock \doi{10.1109/TCE.2007.4429280}.

\bibitem[Gonzalez and Woods(2008)]{gonzalez}
Rafael~C. Gonzalez and Richard~E. Woods.
\newblock \emph{Digital image processing}.
\newblock Prentice Hall, Upper Saddle River, N.J., 2008.
\newblock ISBN 9780131687288 013168728X 9780135052679 013505267X.
\newblock URL
  \url{http://www.amazon.com/Digital-Image-Processing-3rd-Edition/dp/013168728X}.

\bibitem[Huang et~al.(2013)Huang, Cheng, and Chiu]{Huang}
Shih-Chia Huang, Fan-Chieh Cheng, and Yi-Sheng Chiu.
\newblock Efficient contrast enhancement using adaptive gamma correction with
  weighting distribution.
\newblock \emph{IEEE Transactions on Image Processing}, 22\penalty0
  (3):\penalty0 1032--1041, 2013.
\newblock \doi{10.1109/TIP.2012.2226047}.

\bibitem[Al-Ameen et~al.(2015)Al-Ameen, Sulong, Rehman, Al-Dhelaan, Saba, and
  Al-Rodhaan]{Al}
Zohair Al-Ameen, Ghazali Sulong, Amjad Rehman, Abdullah Al-Dhelaan, Tanzila
  Saba, and Mznah Al-Rodhaan.
\newblock An innovative technique for contrast enhancement of computed
  tomography images using normalized gamma-corrected contrast-limited adaptive
  histogram equalization.
\newblock \emph{EURASIP Journal on Advances in Signal Processing}, 04 2015.

\bibitem[Shi et~al.(2020)Shi, Feng, Zhao, Zhang, and He]{Shi}
Zhenghao Shi, Yaning Feng, Minghua Zhao, Erhu Zhang, and Lifeng He.
\newblock Normalized gamma transformation based contrast limited adaptive
  histogram equalization with color correction for sand-dust image enhancement.
\newblock \emph{IET Image Processing}, 14, 03 2020.
\newblock \doi{10.1049/iet-ipr.2019.0992}.

\bibitem[Xia and Liu(2019)]{Xia}
Haiying Xia and Min Liu.
\newblock Non-uniform illumination image enhancement based on retinex and gamma
  correction.
\newblock \emph{Journal of Physics: Conference Series}, 1213:\penalty0 052072,
  06 2019.
\newblock \doi{10.1088/1742-6596/1213/5/052072}.

\bibitem[Kaisserli and Laleg-Kirati(2014)]{Kaisserli2014}
Zineb Kaisserli and Taous-Meriem Laleg-Kirati.
\newblock Image representation and denoising using squared eigenfunctions of
  schrodinger operator.
\newblock \emph{arXiv preprint arXiv:1409.3720}, 2014.

\bibitem[Chahid et~al.(2018)Chahid, Serrai, Achten, and
  Laleg-Kirati]{Chahid2018}
Abderrazak Chahid, Hacene Serrai, Eric Achten, and Taous-Meriem Laleg-Kirati.
\newblock A new roi-based performance evaluation method for image denoising
  using the squared eigenfunctions of the schrödinger operator.
\newblock \emph{2018 40th Annual International Conference of the IEEE
  Engineering in Medicine and Biology Society (EMBC)}, pages 5579--5582, 2018.
\newblock \doi{10.1109/EMBC.2018.8513615}.

\bibitem[Laleg-Kirati et~al.(2013)Laleg-Kirati, Cr{\'{e}}peau, and
  Sorine]{Laleg-Kirati2013}
Taous~Meriem Laleg-Kirati, Emmanuelle Cr{\'{e}}peau, and Michel Sorine.
\newblock {Semi-classical signal analysis}.
\newblock \emph{Mathematics of Control, Signals, and Systems}, 25\penalty0
  (1):\penalty0 37--61, 2013.
\newblock ISSN 1435568X.

\bibitem[Li et~al.(2021)Li, Piliouras, Poghosyan, AlHameed, and
  Laleg-Kirati]{li}
Peihao Li, Evangelos Piliouras, Vahe Poghosyan, Majed AlHameed, and
  Taous-Meriem Laleg-Kirati.
\newblock Automatic detection of epileptiform eeg discharges based on the
  semi-classical signal analysis (scsa) method.
\newblock In \emph{2021 43rd Annual International Conference of the IEEE
  Engineering in Medicine \& Biology Society (EMBC)}, pages 928--931. IEEE,
  2021.

\bibitem[Piliouras(2020)]{evan}
Evangelos Piliouras.
\newblock Contributions to the semi-classical signal analysis method: The
  arterial stiffness assessment case study.
\newblock Master's thesis, King Abdullah University of Science and Technology
  (KAUST), 2020.
\newblock URL \url{http://hdl.handle.net/10754/662544}.

\bibitem[Laleg-Kirati et~al.(2010)Laleg-Kirati, M{\'e}digue, Papelier, Cottin,
  and {Van De Louw}]{validation}
{Taous Meriem} Laleg-Kirati, Claire M{\'e}digue, Yves Papelier, Fran{\c c}ois
  Cottin, and Andry {Van De Louw}.
\newblock Validation of a semi-classical signal analysis method for stroke
  volume variation assessment: A comparison with the picco technique.
\newblock \emph{Annals of Biomedical Engineering}, 38\penalty0 (12):\penalty0
  3618--3629, December 2010.
\newblock ISSN 0090-6964.
\newblock \doi{10.1007/s10439-010-0118-z}.
\newblock Copyright: Copyright 2018 Elsevier B.V., All rights reserved.

\bibitem[Chahid et~al.(2017)Chahid, Serrai, Achten, and
  Laleg-Kirati]{abderrazak2}
Abderrazak Chahid, Hacene Serrai, Eric Achten, and Taous-Meriem Laleg-Kirati.
\newblock Adaptive method for mri enhancement using squared eigenfunctions of
  the schrödinger operator.
\newblock \emph{2017 IEEE Biomedical Circuits and Systems Conference (BioCAS)},
  pages 1--4, 2017.
\newblock \doi{10.1109/BIOCAS.2017.8325107}.

\bibitem[Garcia et~al.(2022)Garcia, Bahloul, and Laleg-Kirati]{Garcia}
Juan Manuel~Vargas Garcia, Mohamed~A. Bahloul, and Taous-Meriem Laleg-Kirati.
\newblock A multiple linear regression model for carotid-to-femoral pulse wave
  velocity estimation based on schrodinger spectrum characterization.
\newblock In \emph{2022 44th Annual International Conference of the IEEE
  Engineering in Medicine \& Biology Society (EMBC)}, pages 143--147, 2022.
\newblock \doi{10.1109/EMBC48229.2022.9871031}.

\bibitem[Vargas et~al.(2023)Vargas, Bahloul, and Laleg-Kirati]{Vargas}
Juan~M. Vargas, Mohamed~A. Bahloul, and Taous-Meriem Laleg-Kirati.
\newblock A learning-based image processing approach for pulse wave velocity
  estimation using spectrogram from peripheral pulse wave signals: An in silico
  study.
\newblock \emph{Frontiers in Physiology}, 14, 2023.
\newblock ISSN 1664-042X.
\newblock \doi{10.3389/fphys.2023.1100570}.
\newblock URL
  \url{https://www.frontiersin.org/articles/10.3389/fphys.2023.1100570}.

\bibitem[Helffer and Laleg‐Kirati(2018)]{HelLal10}
B.~Helffer and T.~M. Laleg‐Kirati.
\newblock On semi‐classical questions related to signal analysis.
\newblock \emph{2018 4th International Conference on Science and Technology
  (ICST)}, pages 1--6, 2018.
\newblock \doi{10.1109/ICSTC.2018.8528579}.
\newblock URL \url{https://ieeexplore.ieee.org/document/8528579}.

\bibitem[Jeon et~al.(2022)Jeon, Park, and Eom]{Jeon}
Jong-Ju Jeon, Tae-Hee Park, and Il-Kyu Eom.
\newblock Sand-dust image enhancement using chromatic variance consistency and
  gamma correction-based dehazing.
\newblock \emph{Sensors}, 22\penalty0 (23), 2022.
\newblock ISSN 1424-8220.
\newblock \doi{10.3390/s22239048}.
\newblock URL \url{https://www.mdpi.com/1424-8220/22/23/9048}.

\bibitem[Park and Eom(2021)]{Park}
Tae~Hee Park and Il~Kyu Eom.
\newblock Sand-dust image enhancement using successive color balance with
  coincident chromatic histogram.
\newblock \emph{IEEE Access}, 9:\penalty0 19749--19760, 2021.
\newblock \doi{10.1109/ACCESS.2021.3054899}.

\bibitem[Sarkar et~al.(2021)Sarkar, Mandal, and Halder]{sarkar}
Kanishka Sarkar, Ardhendu Mandal, and Tanmoy Halder.
\newblock Adaptive power-law and cdf based geometric transformation for low
  contrast image enhancement.
\newblock \emph{Multimedia Tools and Applications}, 80, 02 2021.
\newblock \doi{10.1007/s11042-020-10004-6}.

\bibitem[Arici et~al.(2009)Arici, Dikbas, and Altunbasak]{Arici}
Tarik Arici, Salih Dikbas, and Yucel Altunbasak.
\newblock A histogram modification framework and its application for image
  contrast enhancement.
\newblock \emph{IEEE Transactions on Image Processing}, 18\penalty0
  (9):\penalty0 1921--1935, 2009.
\newblock \doi{10.1109/TIP.2009.2021548}.

\bibitem[Kuran and Kuran(2021)]{Kuran}
Umut Kuran and Emre~Can Kuran.
\newblock Parameter selection for clahe using multi-objective cuckoo search
  algorithm for image contrast enhancement.
\newblock \emph{Intelligent Systems with Applications}, 12:\penalty0 200051,
  2021.
\newblock ISSN 2667-3053.
\newblock \doi{https://doi.org/10.1016/j.iswa.2021.200051}.
\newblock URL
  \url{https://www.sciencedirect.com/science/article/pii/S2667305321000405}.

\bibitem[Kaganami et~al.(2009)Kaganami, Beiji, and Soliman]{Kaganami}
Hassana Kaganami, Zou Beiji, and Mahmoud Soliman.
\newblock Advanced color images enhancement using wavelet and k-means
  clustering.
\newblock \emph{International Journal of Digital Content Technology and its
  Applications}, 5:\penalty0 648--652, 09 2009.
\newblock \doi{10.1109/IIH-MSP.2009.14}.

\bibitem[Lee et~al.(2011)Lee, Park, and Chang]{Lee}
Ji~Lee, Rae-Hong Park, and Soonkeun Chang.
\newblock Local tone mapping using k-means algorithm and automatic gamma
  setting.
\newblock \emph{Consumer Electronics, IEEE Transactions on}, 57:\penalty0 209
  -- 217, 03 2011.

\bibitem[Rousseeuw(1987)]{Rouus}
Peter~J. Rousseeuw.
\newblock Silhouettes: A graphical aid to the interpretation and validation of
  cluster analysis.
\newblock \emph{Journal of Computational and Applied Mathematics}, 20:\penalty0
  53--65, 1987.
\newblock ISSN 0377-0427.
\newblock \doi{https://doi.org/10.1016/0377-0427(87)90125-7}.
\newblock URL
  \url{https://www.sciencedirect.com/science/article/pii/0377042787901257}.

\bibitem[Yang and Deb(2013)]{YANG}
Xin-She Yang and Suash Deb.
\newblock Multiobjective cuckoo search for design optimization.
\newblock \emph{Computers and Operations Research}, 40\penalty0 (6):\penalty0
  1616--1624, 2013.
\newblock ISSN 0305-0548.
\newblock \doi{https://doi.org/10.1016/j.cor.2011.09.026}.
\newblock URL
  \url{https://www.sciencedirect.com/science/article/pii/S0305054811002905}.
\newblock Emergent Nature Inspired Algorithms for Multi-Objective Optimization.

\bibitem[Verma et~al.(2021)Verma, Pant, and Snasel]{Verma}
Shanu Verma, Millie Pant, and Vaclav Snasel.
\newblock A comprehensive review on nsga-ii for multi-objective combinatorial
  optimization problems.
\newblock \emph{IEEE Access}, 9:\penalty0 57757--57791, 2021.
\newblock \doi{10.1109/ACCESS.2021.3070634}.

\bibitem[Moré et~al.(2015)Moré, Brizuela, Ayala, Pinto-Roa, and
  Noguera]{Luis}
Luis~G. Moré, Marcos~A. Brizuela, Horacio~Legal Ayala, Diego~P. Pinto-Roa, and
  José Luis~Vazquez Noguera.
\newblock Parameter tuning of clahe based on multi-objective optimization to
  achieve different contrast levels in medical images.
\newblock In \emph{2015 IEEE International Conference on Image Processing
  (ICIP)}, pages 4644--4648, 2015.
\newblock \doi{10.1109/ICIP.2015.7351687}.

\bibitem[Vijayalakshmi and Nath(2022{\natexlab{a}})]{VIJA}
D.~Vijayalakshmi and Malaya~Kumar Nath.
\newblock A novel multilevel framework based contrast enhancement for uniform
  and non-uniform background images using a suitable histogram equalization.
\newblock \emph{Digital Signal Processing}, 127:\penalty0 103532,
  2022{\natexlab{a}}.
\newblock ISSN 1051-2004.
\newblock \doi{https://doi.org/10.1016/j.dsp.2022.103532}.
\newblock URL
  \url{https://www.sciencedirect.com/science/article/pii/S105120042200149X}.

\bibitem[Abouhawwash et~al.(2017)Abouhawwash, Seada, and Deb]{ABOU}
Mohamed Abouhawwash, Haitham Seada, and Kalyanmoy Deb.
\newblock Towards faster convergence of evolutionary multi-criterion
  optimization algorithms using karush kuhn tucker optimality based local
  search.
\newblock \emph{Computers and Operations Research}, 79:\penalty0 331--346,
  2017.
\newblock ISSN 0305-0548.
\newblock \doi{https://doi.org/10.1016/j.cor.2016.04.026}.
\newblock URL
  \url{https://www.sciencedirect.com/science/article/pii/S0305054816300971}.

\bibitem[Mukhopadhyay et~al.(2022)Mukhopadhyay, Hossain, Malakar, Cuevas, and
  Sarkar]{Mukho}
Souradeep Mukhopadhyay, Sabbir Hossain, Samir Malakar, Erik Cuevas, and Ram
  Sarkar.
\newblock Image contrast improvement through a metaheuristic scheme.
\newblock \emph{Soft Computing}, 07 2022.
\newblock \doi{10.1007/s00500-022-07291-6}.

\bibitem[Xue et~al.(2014)Xue, Zhang, Mou, and Bovik]{Xue}
Wufeng Xue, Lei Zhang, Xuanqin Mou, and Alan~C. Bovik.
\newblock Gradient magnitude similarity deviation: A highly efficient
  perceptual image quality index.
\newblock \emph{IEEE Transactions on Image Processing}, 23\penalty0
  (2):\penalty0 684--695, 2014.
\newblock \doi{10.1109/TIP.2013.2293423}.

\bibitem[Wang et~al.(2015)Wang, Ma, Yeganeh, Wang, and Lin]{Wang}
Shiqi Wang, Kede Ma, Hojatollah Yeganeh, Zhou Wang, and Weisi Lin.
\newblock A patch-structure representation method for quality assessment of
  contrast changed images.
\newblock \emph{IEEE Signal Processing Letters}, 22\penalty0 (12):\penalty0
  2387--2390, 2015.
\newblock \doi{10.1109/LSP.2015.2487369}.

\bibitem[Vijayalakshmi and Nath(2022{\natexlab{b}})]{Vijaya}
D.~Vijayalakshmi and Malaya~Kumar Nath.
\newblock A novel multilevel framework based contrast enhancement for uniform
  and non-uniform background images using a suitable histogram equalization.
\newblock \emph{Digital Signal Processing}, 127:\penalty0 103532,
  2022{\natexlab{b}}.
\newblock ISSN 1051-2004.
\newblock \doi{https://doi.org/10.1016/j.dsp.2022.103532}.
\newblock URL
  \url{https://www.sciencedirect.com/science/article/pii/S105120042200149X}.

\bibitem[Kumar and Bhandari(2022)]{Kumar}
Reman Kumar and Ashish~Kumar Bhandari.
\newblock Fuzzified contrast enhancement for nearly invisible images.
\newblock \emph{IEEE Transactions on Circuits and Systems for Video
  Technology}, 32\penalty0 (5):\penalty0 2802--2813, 2022.
\newblock \doi{10.1109/TCSVT.2021.3098763}.

\bibitem[Singh et~al.(2022)Singh, Bhandari, and Kumar]{Pallavi}
Pallavi Singh, Ashish~Kumar Bhandari, and Reman Kumar.
\newblock Naturalness balance contrast enhancement using adaptive gamma with
  cumulative histogram and median filtering.
\newblock \emph{Optik}, 251:\penalty0 168251, 2022.
\newblock ISSN 0030-4026.
\newblock \doi{https://doi.org/10.1016/j.ijleo.2021.168251}.
\newblock URL
  \url{https://www.sciencedirect.com/science/article/pii/S0030402621017745}.

\bibitem[Dixit et~al.(2021)Dixit, Yadav, and Mishra]{Dixit}
Avadhesh~Kumar Dixit, Rakesh~Kumar Yadav, and Ramapati Mishra.
\newblock Image contrast optimization using local color correction and fuzzy
  intensification.
\newblock \emph{International Journal of Advanced Computer Science and
  Applications}, 12, 2021.

\bibitem[Asamoah et~al.(2018)Asamoah, Oppong, Oppong, and Danso]{Asamoah2018}
Dominic Asamoah, Emmanuel Ofori Oppong, Stephen Opoku Oppong, and Juliana Danso.
\newblock Measuring the Performance of Image Contrast Enhancement Technique.
\newblock \emph{International Journal of Computer Applications}, volume 181, number 22, pages 6--13, October 2018.
\newblock \doi{10.5120/ijca2018917899}.
\newblock URL \url{http://www.ijcaonline.org/archives/volume181/number22/30015-2018917899}.

\bibitem[Attar et~al.(2018)Attar, Bhattacharjee, Kumar, and Dey]{attar}
Tushar Attar, Tushar Bhattacharjee, Rajesh Kumar, and Nilanjan Dey.
\newblock Parametric Optimization of Logarithmic Transformation using GWO for Enhancement and Denoising of MRI Images.
\newblock \emph{2018 3rd International Conference and Workshops on Recent Advances and Innovations in Engineering (ICRAIE)}, pages 1--7, 2018.
\newblock \doi{10.1109/ICRAIE.2018.8710393}.

\bibitem[Moreno López et~al.(2021)Moreno López, Frederick, and Ventura]{MORENO}
Marc Moreno López, Joshua M. Frederick, and Jonathan Ventura.
\newblock Evaluation of MRI Denoising Methods Using Unsupervised Learning.
\newblock \emph{Frontiers in Artificial Intelligence}, volume 4, 2021.
\newblock \doi{10.3389/frai.2021.642731}.
\newblock URL \url{https://www.frontiersin.org/articles/10.3389/frai.2021.642731}.

\bibitem[Sahnoun et~al.(2018)Sahnoun, Kallel, Dammak, Mhiri, Ben Mahfoudh, and Ben Hamida]{sahnon2018}
Mouna Sahnoun, Fathi Kallel, Mariem Dammak, Chokri Mhiri, Kheireddine Ben Mahfoudh, and Ahmed Ben Hamida.
\newblock A comparative study of MRI contrast enhancement techniques based on Traditional Gamma Correction and Adaptive Gamma Correction: Case of multiple sclerosis pathology.
\newblock \emph{2018 4th International Conference on Advanced Technologies for Signal and Image Processing (ATSIP)}, pages 1--7, 2018.
\newblock \doi{10.1109/ATSIP.2018.8364467}.

\bibitem[Mirza et~al.(2022)Mirza, Siddiq, and Khan]{Mirza}
Muhammad Mirza, Asif Siddiq, and Ishtiaq Khan.
\newblock A comparative study of medical image enhancement algorithms and quality assessment metrics on COVID-19 CT images.
\newblock \emph{Signal, Image and Video Processing}, volume 17, 2022.
\newblock \doi{10.1007/s11760-022-02214-2}.

\bibitem[Zhou et~al.(2019)Zhou, Shi, Lai, and Jimenez]{Zhou2019}
Yuanping Zhou, Changqin Shi, Bingyan Lai, and Giorgos Jimenez.
\newblock Contrast enhancement of medical images using a new version of the World Cup Optimization algorithm.
\newblock \emph{Quantitative Imaging in Medicine and Surgery}, volume 9, issue 9, pages 1528--1547, 2019.
\newblock \doi{10.21037/qims.2019.08.19}.

\bibitem[Singh et~al.(2018)Singh, Verma, and Sharma]{Singh2018}
M. Singh, A. Verma, and N. Sharma.
\newblock An Optimized Cascaded Stochastic Resonance for the Enhancement of Brain MRI.
\newblock \emph{IRBM}, volume 39, issue 5, pages 334--342, November 2018.
\newblock \doi{10.1016/j.irbm.2018.08.002}.

\bibitem[Rahman and Paul(2023)Rahman and Paul]{tripa}
Hafijur Rahman and Gour Chandra Paul.
\newblock Tripartite sub-image histogram equalization for slightly low contrast gray-tone image enhancement.
\newblock \emph{Pattern Recognition}, volume 134, page 109043, 2023.
\newblock \doi{https://doi.org/10.1016/j.patcog.2022.109043}.
\newblock URL \url{https://www.sciencedirect.com/science/article/pii/S0031320322005234}.

\bibitem[Cao et~al.(2022)Cao, Huang, Tian, Huang, Wang, and Zhi]{IAGCWD}
Gang Cao, Lihui Huang, Huawei Tian, Xianglin Huang, Yongbin Wang, and Ruicong Zhi.
\newblock Contrast Enhancement of Brightness-Distorted Images by Improved Adaptive Gamma Correction.
\newblock \emph{arXiv preprint}, 2022.
\newblock URL \url{https://arxiv.org/abs/1709.04427}.

\bibitem[Li et~al.(2024)Li, Liu, and Ling]{PWGC}
Xiangsheng Li, Manlu Liu, and Qiang Ling.
\newblock Pixel-Wise Gamma Correction Mapping for Low-Light Image Enhancement.
\newblock \emph{IEEE Transactions on Circuits and Systems for Video Technology}, volume 34, number 2, pages 681--694, 2024.
\newblock \doi{10.1109/TCSVT.2023.3286802}.

\bibitem[Zhao et~al.(2023)Zhao, Huang, Che, Xie, Liu, and Wang]{c-fuzzy1}
Yunlan Zhao, Zhiyong Huang, Hangjun Che, Fang Xie, Man Liu, and Mengyao Wang.
\newblock Segmentation of Brain Tissues from MRI Images Using Multitask Fuzzy Clustering Algorithm.
\newblock \emph{Journal of Healthcare Engineering}, volume 2023, pages 1--15, February 2023.
\newblock \doi{10.1155/2023/4387134}.

\bibitem[Srivastava et~al.(2023)Srivastava, Vidyarthi, and Jain]{fuzzy-fet}
Somya Srivastava, Ankit Vidyarthi, and Shikha Jain.
\newblock Fetal Head Segmentation Using Optimized K-Means and Fuzzy C-Means on 2D Ultrasound Images.
\newblock In Abhishek Swaroop, Vineet Kansal, Giancarlo Fortino, and Aboul Ella Hassanien, editors, \emph{Proceedings of Fourth Doctoral Symposium on Computational Intelligence}, pages 441--453, Singapore, 2023. Springer Nature Singapore.
\newblock ISBN 978-981-99-3716-5.

\bibitem[Pan et~al.(2022)Pan, Zheng, and Jeon]{Pan}
Xiaoyan Pan, Yuhui Zheng, and Byeungwoo Jeon.
\newblock Robust Segmentation Based on Salient Region Detection Coupled Gaussian Mixture Model.
\newblock \emph{Information}, volume 13, page 98, February 2022.
\newblock \doi{10.3390/info13020098}.

\bibitem[Hum et~al.(2022)Hum, Tee, Yap, Mokayed, Tan, Salim, and Lai]{Unconds}
Yan Hum, Yee Tee, Wun-She Yap, Hamam Mokayed, Tian Swee Tan, Maheza Salim, and Khin Wee Lai.
\newblock A contrast enhancement framework under uncontrolled environments based on just noticeable difference.
\newblock \emph{Signal Processing: Image Communication}, volume 103, page 116657, February 2022.
\newblock \doi{10.1016/j.image.2022.116657}.

\bibitem[Dinh(2023)Dinh]{medical}
Phu-Hung Dinh.
\newblock Medical image fusion based on enhanced three-layer image decomposition and Chameleon swarm algorithm.
\newblock \emph{Biomedical Signal Processing and Control}, volume 84, page 104740, February 2023.
\newblock \doi{10.1016/j.bspc.2023.104740}.

\bibitem[Rahman and Paul(2023)Rahman and Paul]{RAHMAN}
Hafijur Rahman and Gour Chandra Paul.
\newblock Tripartite sub-image histogram equalization for slightly low contrast gray-tone image enhancement.
\newblock \emph{Pattern Recognition}, volume 134, page 109043, 2023.
\newblock \doi{https://doi.org/10.1016/j.patcog.2022.109043}.
\newblock URL \url{https://www.sciencedirect.com/science/article/pii/S0031320322005234}.

\bibitem[Zhang et~al.(2023)Zhang, Jin, Zhuang, Liang, and Li]{underW}
Weidong Zhang, Songlin Jin, Peixian Zhuang, Zheng Liang, and Chongyi Li.
\newblock Underwater Image Enhancement via Piecewise Color Correction and Dual Prior Optimized Contrast Enhancement.
\newblock \emph{IEEE Signal Processing Letters}, volume 30, pages 229--233, 2023.
\newblock \doi{10.1109/LSP.2023.3255005}.

\bibitem[Rasheed et~al.(2023)Rasheed, Shi, and Khan]{revlli}
Muhammad Rasheed, Daming Shi, and Hufsa Khan.
\newblock A comprehensive experiment-based review of low-light image enhancement methods and benchmarking low-light image quality assessment.
\newblock \emph{Signal Processing}, volume 204, page 108821, March 2023.
\newblock \doi{10.1016/j.sigpro.2022.108821}.

\end{thebibliography}

\end{document}